\newcommand{\model}{\textsf{DualFair}}
\newcommand{\tabmix}{\textsf{TabMix}}

\documentclass[sigconf]{acmart}
\AtBeginDocument{%
  \providecommand\BibTeX{{%
    \normalfont B\kern-0.5em{\scshape i\kern-0.25em b}\kern-0.8em\TeX}}}

\setcopyright{acmcopyright}
\copyrightyear{2023}
\acmYear{2023}
\setcopyright{rightsretained}
\acmConference[WWW '23]{Proceedings of the ACM Web Conference 2023}{April
30-May 4, 2023}{Austin, TX, USA}
\acmBooktitle{Proceedings of the ACM Web Conference 2023 (WWW '23), April
30-May 4, 2023, Austin, TX, USA}
\acmDOI{10.1145/3543507.3583480}
\acmISBN{978-1-4503-9416-1/23/04}

\acmConference[WWW '23] {Proceedings of the ACM Web Conference 2023}{April 30--May 4, 2023}{Austin, Texas, USA.}

\usepackage{subcaption}
\usepackage[ruled,linesnumbered]{algorithm2e}
\usepackage{enumitem}
\usepackage{multirow}
\usepackage{balance}

\begin{document}


\title[DualFair]{DualFair: Fair Representation Learning at Both Group and Individual Levels via Contrastive Self-supervision}


\author{Sungwon Han}
\orcid{0000-0002-1129-760X}
\affiliation{%
  \institution{KAIST}
  \city{Daejeon}
  \country{South Korea}
}
\email{lion4151@kaist.ac.kr}

\author{Seungeon Lee}
\orcid{0000-0002-9756-0068}
\affiliation{%
  \institution{KAIST}
  \city{Daejeon}
  \country{South Korea}
}
\email{archon159@kaist.ac.kr}

\author{Fangzhao Wu}
\orcid{0000-0001-9138-1272}
\affiliation{%
  \institution{Microsoft Research Asia}
  \city{Beijing}
  \country{China}
}
\email{wufangzhao@gmail.com}

\author{Sundong Kim}
\orcid{0000-0001-9687-2409}
\affiliation{%
 \institution{GIST}
 \city{Gwangju}
 \country{South Korea}
 }
 \email{sundong@gist.ac.kr}

\author{Chuhan Wu}
\orcid{0000-0001-5730-8792}
\affiliation{%
  \institution{Tsinghua University}
  \city{Beijing}
  \country{China}
}
\email{wuchuhan15@gmail.com}

\author{Xiting Wang}
\orcid{0000-0001-5768-1095}
\affiliation{%
  \institution{Microsoft Research Asia}
  \city{Beijing}
  \country{China}
}
\email{xitwan@microsoft.com}

\author{Xing Xie}
\orcid{0000-0002-8608-8482}
\affiliation{%
  \institution{Microsoft Research Asia}
  \city{Beijing}
  \country{China}
}
\email{xingx@microsoft.com}

\author{Meeyoung Cha}
\orcid{0000-0003-4085-9648}
\affiliation{%
 \institution{IBS \& KAIST}
 \city{Daejeon}
 \country{South Korea}
 }
 \email{mcha@ibs.re.kr}

\renewcommand{\shortauthors}{Anon., et al.}

\begin{abstract}
Algorithmic fairness has become an important machine learning problem, especially for mission-critical Web applications.
This work presents a self-supervised model, called \model{}, that can debias sensitive attributes like gender and race from learned representations.
Unlike existing models that target a single type of fairness, our model jointly optimizes for two fairness criteria---\textit{group fairness} and \textit{counterfactual fairness}---and hence makes fairer predictions at both the group and individual levels.
Our model uses contrastive loss to generate embeddings that are indistinguishable for each protected group, while forcing the embeddings of counterfactual pairs to be similar. 
It then uses a self-knowledge distillation method to maintain the quality of representation for the downstream tasks. 
Extensive analysis over multiple datasets confirms the model's validity and further shows the synergy of jointly addressing two fairness criteria, suggesting the model's potential value in fair intelligent Web applications.
\looseness=-1
\end{abstract}



\maketitle

\section{Introduction}

Machine learning techniques are being used in many mission-critical real-world Web applications, such as recommendation~\cite{an2019neural}, hiring~\cite{hoffman2018discretion}, and advertisement of profiles~\cite{sweeney2013discrimination}. Many tasks are subject to stereotypes or societal biases embedded in data, leading models to treat individuals of particular attributes unfairly. Notable examples include facial recognition models that fail to recognize people with darker skin~\cite{buolamwini2018gender} or recruiting models that favor men over women candidates with comparable work experience~\cite{kiritchenko2018examining}. Lacking fairness in these tasks aggravates social inequity and even harms individuals and society. As a response, a growing number of studies are dedicated to algorithmic fairness~\cite{dwork2012fairness,wu2021fairness}.

Fairness is also important in representation learning. Fair representation learning is useful when data needs to be shared without any prior knowledge of the downstream task. 
Previous research has focused on finding data representations that apply to diverse tasks while considering fairness. For example, LAFTR~\cite{madras2018learning} and VFAE~\cite{louizos2016variational} introduce an adversarial concept, and studies like~\cite{song2019learning} and~\cite{tsai2021conditional} maximize the conditional mutual information with fairness constraints to reduce the potential bias. These approaches generate fairer embeddings, albeit at the cost of performance degradation in downstream tasks. They also consider only group-level fairness, missing out on individual-level fairness~\cite{binns2020apparent}.

We propose \model{}, a self-supervised learning method that debiases sensitive information through fairness-aware contrastive learning while preserving rich expressiveness (i.e., representation quality) through self-knowledge distillation. 
\model{} can learn data representations satisfying two types of fairness, i.e., \textit{group fairness} and 
\textit{counterfactual fairness}.
The former (a.k.a. demographic parity) requires every protected group be treated in the same way as any advantaged group, and the latter requires the model to treat individuals in a counterfactual relationship (i.e., those who share similar traits except for the sensitive attribute) alike~\cite{kusner2017counterfactual}. 
Counterfactual fairness is a type of fairness defined at the individual level. It removes bias from sensitive information by comparing against synthetic individuals from the counterfactual world. \looseness=-1

\begin{figure}[t!]
\centering
\begin{subfigure}[t]{0.40\textwidth}
      \centering\includegraphics[width=0.9\textwidth]{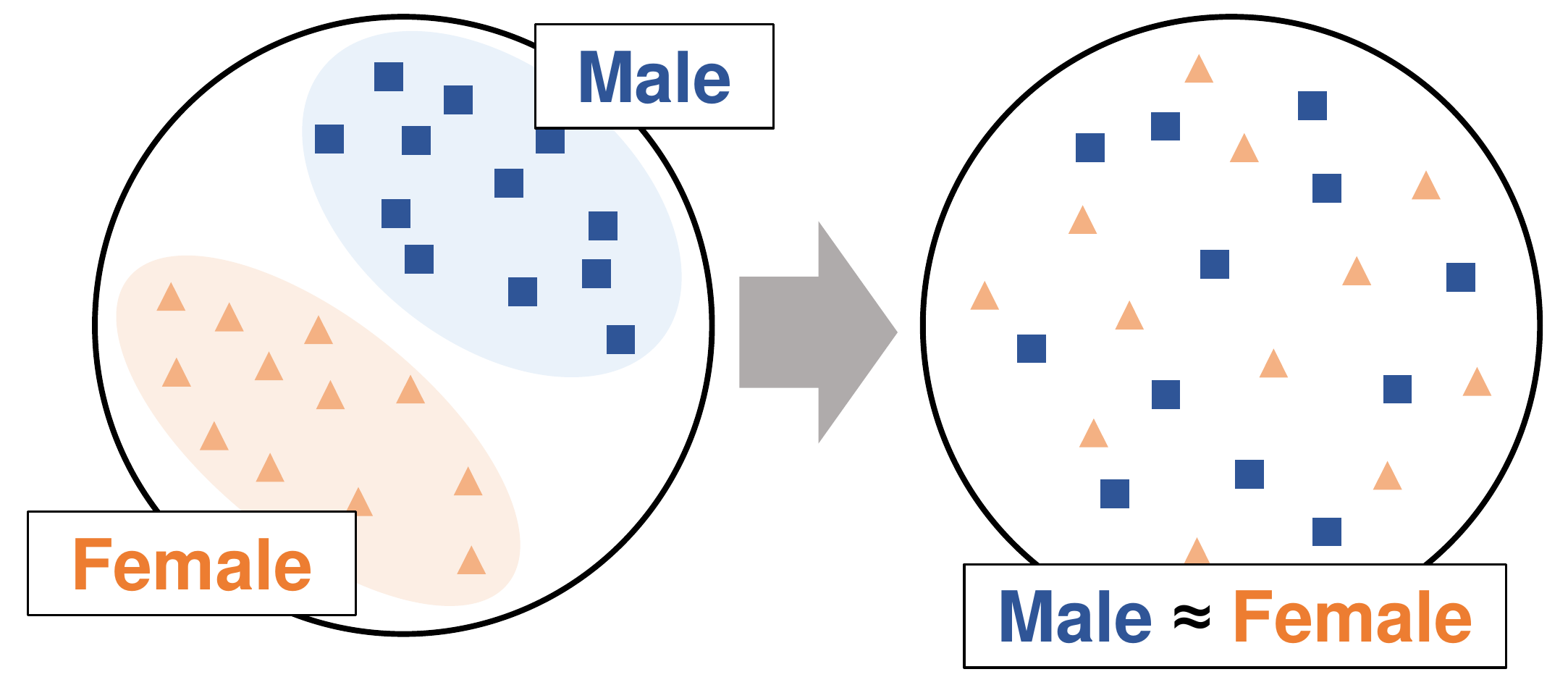}
      \subcaption{Group fairness: Embeddings of different member groups (e.g., by gender) should not be distinguished. This makes unbiased predictions possible at the group level.}
      \label{fig:intro1}
      \vspace{1mm}
\end{subfigure}
\begin{subfigure}[t]{0.40\textwidth}
      \centering\includegraphics[width=1\textwidth]{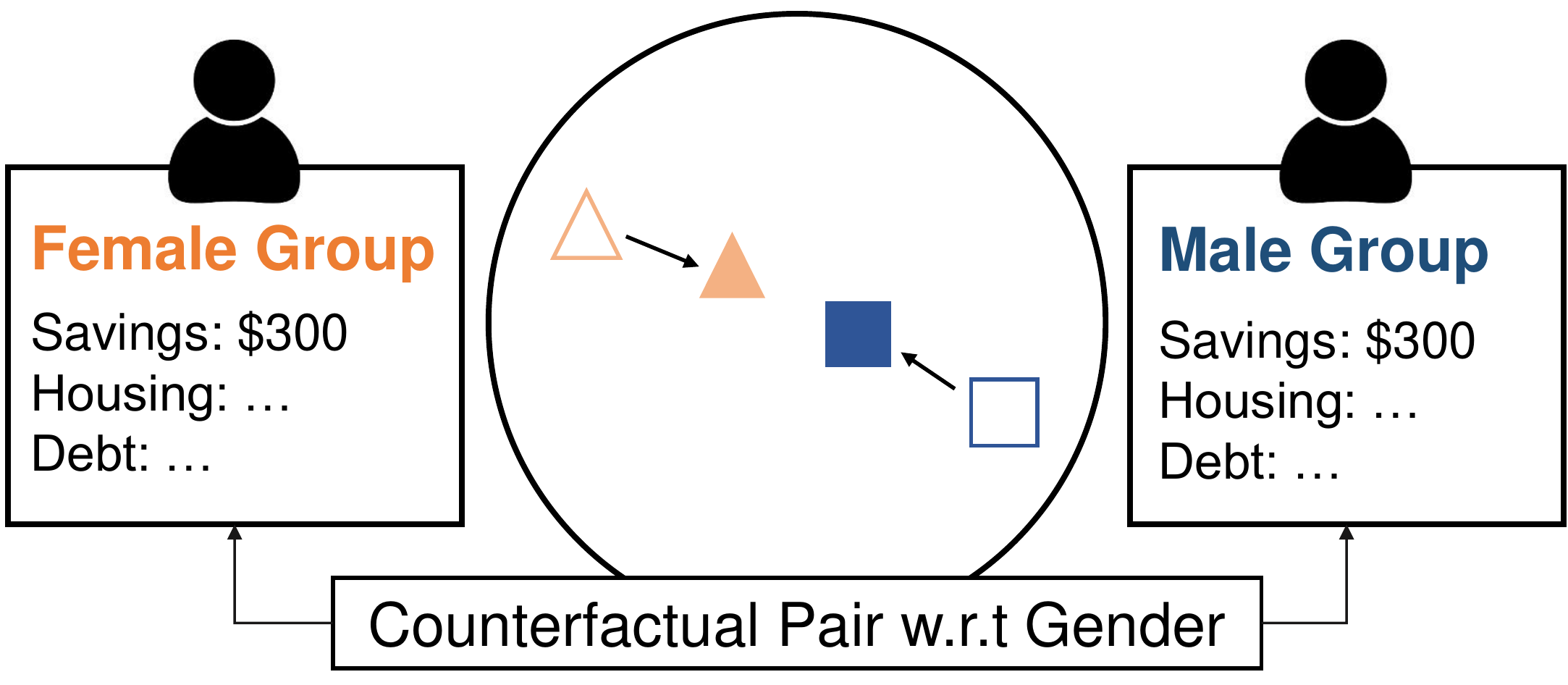} 
      \subcaption{Counterfactual fairness (i.e., individual-level fairness): Individuals of similar traits should have similar embeddings irrespective of their sensitive attributes. This makes unbiased predictions possible at the individual level.}
      \label{fig:intro2}
      \vspace{-1mm}
\end{subfigure}
\caption{Illustration of two fairness criteria. 
\vspace{-3mm}} 
\label{fig:motivate}
\end{figure}

Because the model is unaware of downstream tasks during training, adding fairness-related regularization terms (e.g., minimizing demographic parity from model predictions~\cite{kamishima2012fairness}) or fairness constraints (e.g., limiting the prediction difference among groups below the threshold~\cite{zafar2019fairness}) as in other research is not feasible. Instead, we propose the following alternative goals for our loss design, which are illustrated in Figure~\ref{fig:motivate}:
\begin{enumerate}
\item
\textbf{\textsf{Group fairness}}: Data points from every sensitive group have the same distribution across the embedding space; thus, their group membership cannot be identified. 

\item \textbf{\textsf{Counterfactual fairness}}: 
Data points from those in a counterfactual relationship are located close in the embedding space. This means the embedding distance between an individual and its counterfactual version should be minimized.
\looseness=-1 
\end{enumerate}
These goals can apply universally to any downstream task.
By making sensitive attributes indistinguishable in the embedding space, a downstream classifier cannot determine which data point belongs to the protected group and consequently produce unbiased predictions. The same applies to embeddings of counterfactual pairs.\looseness=-1

To implement these fairness objectives, we first propose a cyclic variational autoencoder (C-VAE) model to generate counterfactual samples (Sec.~3.2). This sample generation task is non-trivial due to the high correlation between data features and sensitive attributes. Next is to create fairness-aware embeddings from the input data. We employ contrastive learning, which learns representations based on the similarity between instances by modifying the selection of positive and negative samples. We select negative samples from the same protected group and select positive samples from counterfactual versions (Sec.~3.3).
In addition, we use self-knowledge distillation to extract semantic features and enforce consistent embeddings between the original and its augmentation to maintain a high representation quality (Sec.~3.4). 
Experiments demonstrate that the proposed framework can generate data embeddings that satisfy both fairness criteria while preserving prediction accuracy for a variety of downstream tasks. 
The main contributions of this paper are summarized below. \smallskip \looseness=-1
\begin{itemize}[nosep,leftmargin=1em,labelwidth=*,align=left]
    \item We propose a \textbf{self-supervised representation learning framework (\model{})} that simultaneously debiases sensitive attributes at both group and individual levels.
    
    \item We introduce the C-VAE model to generate counterfactual samples and propose \textbf{fairness-aware contrastive loss} to meet the two fairness criteria jointly.
    
    \item We design the \textbf{self-knowledge distillation loss} to maintain representation quality by minimizing the embedding discrepancy between original and perturbed instances.
    
    \item Experimental results on six real-world datasets confirm that \model{} generates a fair embedding for sensitive attributes while maintaining high representation quality. The ablation study further shows a synergistic effect of the two fairness criteria.
\end{itemize}
Codes are released at a GitHub repository.\footnote{https://github.com/Sungwon-Han/DualFair} 

\section{Related Works}

\subsection{Fairness in Machine Learning}
\textit{Fairness} is a conceptual term, and cannot be measured in a straightforward manner. Instead, there are several criteria to observe it from different perspectives: unawareness, group fairness, and individual fairness~\cite{gajane2017formalizing,verma2018fairness}. Removing sensitive attributes from data is a simple way to achieve fairness through unawareness~\cite{chen2019fairness}.
However, it can be brittle if some hidden features are highly correlated with sensitive attributes. Group fairness states that subjects in different groups (e.g., gender) should have an equal probability of being assigned to the predicted class~\cite{conitzer2019group,gajane2017formalizing}. Demographic parity and equalized odds are two measures of group fairness~\cite{hardt2016equality,zafar2017fairness}. Individual fairness is a fine-grained criterion that treats similar individuals as similarly as possible ~\cite{dwork2012innovations}. Counterfactual fairness is one alternative to individual fairness that assumes a counterfactual sample by flipping the sensitive attributes and treats it similarly to the original one~\cite{russel2017when}. \looseness=-1

Researchers also focus on diverse learning steps to ensure fairness. Elazar et al.~\cite{elazar2018advesarial} focus on fairness of the input dataset by noting that  a fair decision is made with the model trained with an unbiased dataset. However, Wang et al.~\cite{wang2019balanced} demonstrate that input-wise fairness does not entirely support fair decisions in large-scale datasets. Therefore, post-processing techniques are proposed to secure fairness, such as hiding the sensitive attribute information from trained representations by null space projection~\cite{ravfogel2020null} or identifying the subspace of sensitive attributes~\cite{bolukbasi2016man}.

With the advent of representation learning, recent self-supervised learning approaches aim to produce a fair representation of individual instances without knowing downstream tasks (i.e., treatment-level fairness)~\cite{kose2021fairness}.
These fair representation learning approaches add fairness-related objectives in training steps, or apply adversarial learning objectives to obtain fair representations~\cite{li2018towards, zhang2018mitigating, barrett2019adversarial, han2021mitigating}. For example, LAFTR~\cite{madras2018learning} employs a discriminator that detects sensitive attribute information, while generators make indistinguishable representations against discriminators. VFAE~\cite{louizos2016variational} proposes a variational autoencoder with regularization using Maximum Mean Discrepancy (MMD) to learn fair representations.

\begin{figure*}[t!]
\centering
\includegraphics[width=0.93\textwidth]{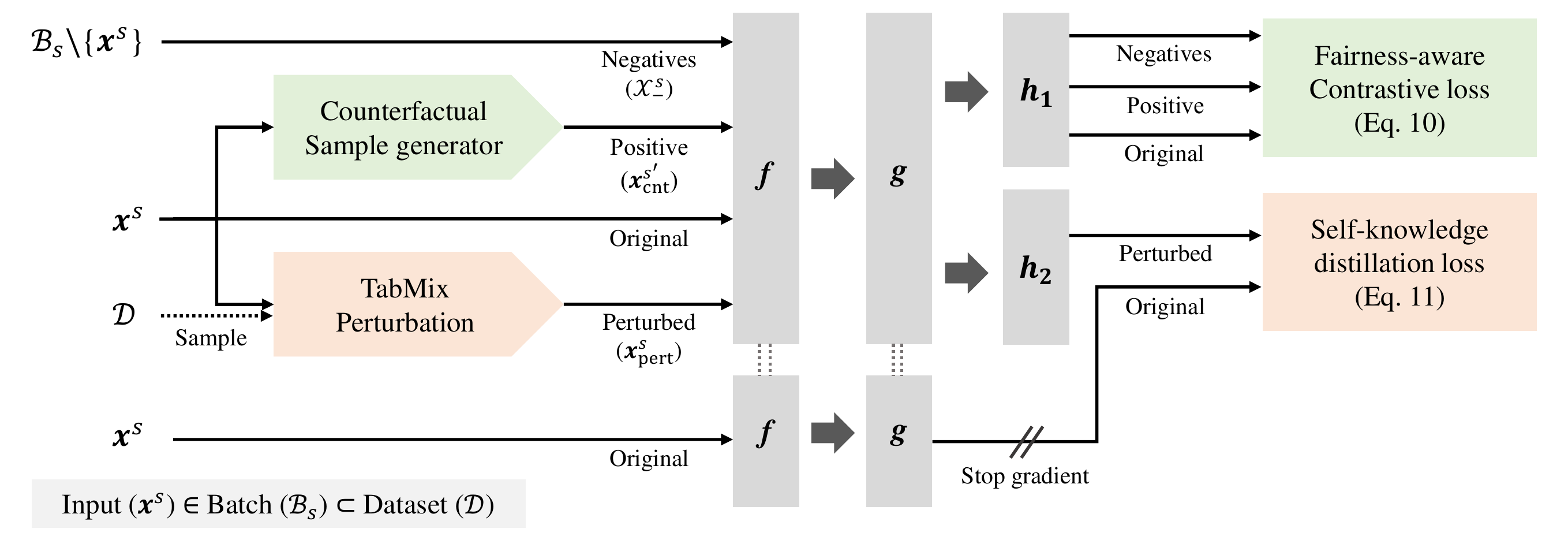}
\caption{The overall architecture of \model{}, where $f$, $g$, ($h_1$, $h_2$) denote the backbone network, the projection head, and two prediction heads, respectively. \model{} aims to achieve both fairness and representation quality by jointly optimizing the fairness-aware contrastive loss and the self-knowledge distillation loss.}
\label{fig:main_training}
\end{figure*}

Song et al.~\cite{song2019learning} proposes a user-centric approach that allows users to control the level of fairness and maximize the expressiveness of representations in the form of conditional mutual information with the preset fairness threshold. Tsai et al.~\cite{tsai2021conditional} aim to optimize the same objective but maximize the lower bound of conditional mutual information via InfoNCE objective instead. However, these methods only consider the single group fairness objective and often lose the expressiveness during the training~\cite{burke2017multisided,burke2018balanced}. Specifically, group fairness helps achieve anti-discrimination for protected groups, but individual justice is not guaranteed~\cite{binns2020apparent}.
Our research objective is to implement multiple algorithmic fairness concepts on a single deep model, a topic that is explored less in the literature. We seek to achieve both group fairness at a coarse-grained level and counterfactual fairness at an individual level while preserving a high level of representational quality. \looseness=-1

\subsection{Contrastive Self-Supervised Learning} \label{ref:2}
The fundamental idea of contrastive learning is to minimize the distance between similar (i.e., positive) instances while maximizing the distance among dissimilar (i.e., negative) instances~\cite{chen2020simple,he2020momentum}. SimCLR~\cite{chen2020simple} utilizes augmented images as positives while the other images in the same batch as negatives. Maintaining a similar contrastive concept, MoCo~\cite{he2020momentum} exploits a momentum encoder and proposes a dynamic dictionary with a queue to handle negative samples efficiently in both performance and memory perspectives. InfoNCE loss~\cite{oord2018representation} is often used in contrastive learning. Minimizing this loss increases mutual information between positive pairs so that the model can extract the consistent features between the original and augmented samples. 

\section{Methodology}
\subsection{Overview}
We present \model{}, a self-supervised learning method that ensures both group and counterfactual fairness criteria while maintaining high representation quality with the design of two special losses. We use a pictorial sketch in Figure~\ref{fig:main_training} to describe them.

First is the \textbf{fairness-aware contrastive loss}, which treats individuals in \textit{counterfactual relationships alike} (i.e., counterfactual fairness) and ensures \textit{non-distinguishable embeddings} over sensitive attributes (i.e., group fairness). A key to this loss is the design of a sample generator that produces a counterfactual version $\mathbf{x}_{\text{cnt}}^{s'}$ of an input item $\mathbf{x}^{s}$ by maintaining its latent characteristics but flipping the sensitive attribute $s \rightarrow s'$ (Sec. 3.2). In a job interview, for instance, this corresponds to a hypothetical decision if the applicant's gender were to change, with all other latent traits, such as education and work experience, remaining unchanged~\cite{kusner2017counterfactual}. Given a dataset $\mathcal{D}$ with sensitive attributes $S$ like gender, the fairness-aware contrastive loss is defined on the batch $\mathcal{B}_{s}$, whose data instances share the same sensitive attribute $s\in S$ (e.g., gender is female). The loss maximizes agreements between the original item and its counterfactual version generated by our generator (i.e., ensuring counterfactual fairness) and minimizes agreements between items with the same sensitive attribute (i.e., ensuring group fairness) (Sec. 3.3). \looseness=-1

\begin{figure*}[t!]
\centering
\begin{minipage}[t]{0.68\textwidth}
       \centering\includegraphics[width=\textwidth]{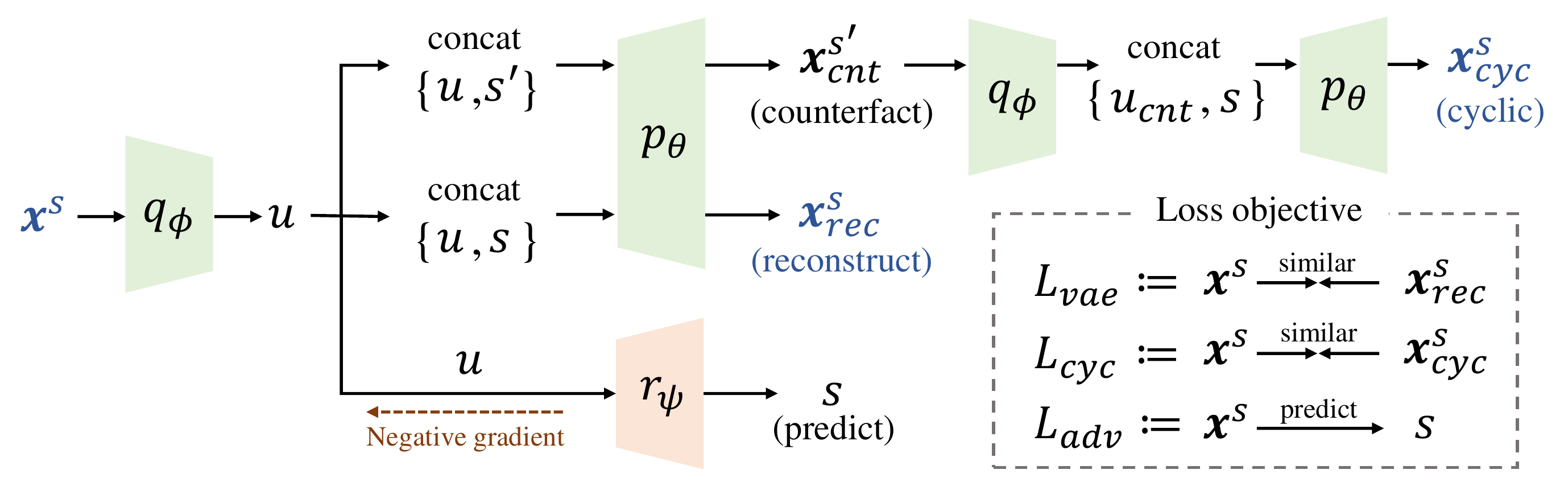}
      \caption{Design of our counterfactual sample generator. Adversarial training over VAE eliminates sensitive information $s$ from $\mathbf{u}$. Cyclic consistency loss makes training more stable by ensuring the double-flipped sample ($\mathbf{x}^s_{\text{cyc}}$) to be same as the original ($\mathbf{x}^s$). \looseness=-1}
      \label{fig:vae}
\end{minipage}
\hspace{5mm}
\begin{minipage}[t]{0.25\textwidth}
       \centering\includegraphics[width=\textwidth]{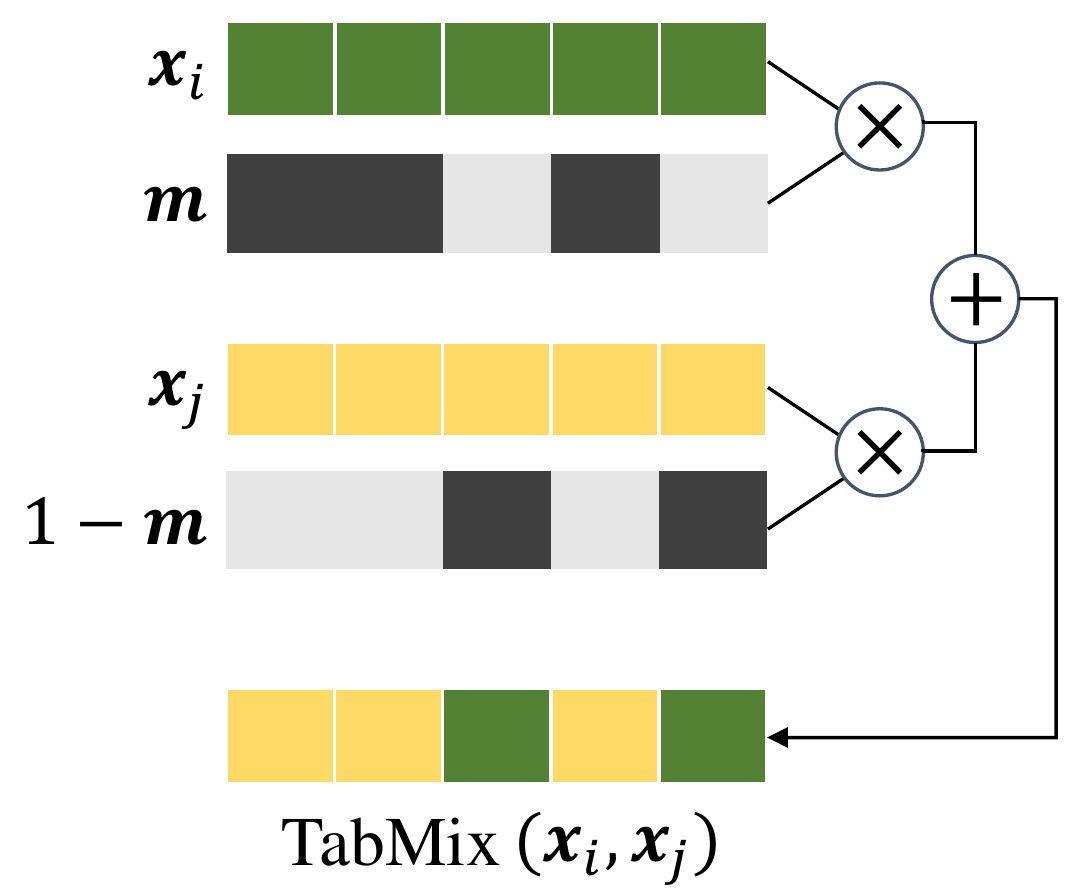}
      \caption{Illustration of \tabmix{} operation for self-knowledge distillation ($m$ denotes a mask). \looseness=-1}
      \label{fig:tabmix}
\end{minipage}
\end{figure*}

Second is the \textbf{self-knowledge distillation loss} for extracting semantic information from the instance and maintaining the performance of downstream tasks (Sec. 3.4). Consider a Siamese network that has two heads: one for the student branch and the other for the teacher branch. We impose the perturbed instance's embedding from the student branch to be similar to that of the original instance from the teacher branch. The rationale behind this treatment is that a perturbation that does not largely deform the original content should not change the learned representation. We introduce our unique perturbation module \tabmix{} for an efficient self-knowledge distillation process. \looseness=-1

\subsection{Counterfactual Sample Generation}
Our sample generator creates data following the definition of a counterfactual relationship in~\cite{kusner2017counterfactual}. Let us denote three variables $V, U, R$, where $V$ is a set of observable variables, including the sensitive attribute $S$ and other attributes $X$. $U$ is a set of latent variables independent of $S$ or $V$. $R$ is a set of functions representing causal relationships from $U$ to $V$, or between $V$. Considering a causal model with ($V, U, R$), counterfactual inference works in three steps: \looseness=-1
\begin{enumerate}[leftmargin=*]
\item Calculate the posterior distribution of latent variables $U$ from the input instance.
\item Choose a target sensitive attribute $s\in S$ (e.g., gender) and reformulate a set of functions $R$ assuming the sensitive attribute $s \in S$ is fixed in the causal graph.
\item Infer $V$ from $U$ and $s \in S$ using re-formulated $R$. 
\end{enumerate}
Extracting the latent variable $U$ is non-trivial because it should explain observable variables sufficiently without revealing information about sensitive attributes.

For this task, we introduce the cyclic variational autoencoder (C-VAE) depicted in Figure~\ref{fig:vae}. Our counterfactual sample generator is trained in the form of $p(\mathbf{x}^s, \mathbf{u}, s) = p(\mathbf{u})p(s)p(\mathbf{x}^s|\mathbf{u}, s)$, where $\mathbf{u}$ and $s$ represent the latent variable and the sensitive attribute in the input $\mathbf{x}^s$. $p(\mathbf{u})$ and $p(s)$ represent the prior distribution of the latent variable $\mathbf{u}$ from the encoder and the sensitive attribute $s$. $p(\mathbf{u})$ is assumed to follow a standard normal distribution~\cite{kingma2013auto}. Then $p(\mathbf{x}^s|\mathbf{u}, s)$ is a likelihood function from the decoder that reconstructs the input sample $\mathbf{x}^s$ from $\mathbf{u}$ and $s$. The encoder and decoder networks are denoted as $q_\phi$ and $p_\theta$. We formulate $L_{\text{vae}}(\mathbf{x}^s)$ from the variational lower bound of $\log p(\mathbf{x})$ as follows: \looseness=-1
\begin{align}
    L_{\text{vae}}(\mathbf{x}^s)=\text{KL}(q_\phi (\mathbf{u}  | \mathbf{x}^s) || p(\mathbf{u}))-\mathbb{E}_{q_\phi (\mathbf{u}|\mathbf{x}^s)}[\log p_\theta (\mathbf{x}^s | \mathbf{u},s)]. \label{eq:vae_loss}
\end{align}
Latent variable $\mathbf{u}$ from the encoder $q_\phi$ and sensitive attribute $s$ are not guaranteed to be independent even if they correspond to $U$ and $S$ in the causal model that are independent. To ensure independence, we introduce an additional adversarial objective and lead the model to eliminate unnecessary sensitive information from $\mathbf{u}$. A discriminator $r_\psi$ is trained with a cross-entropy loss to predict the sensitive attribute $s$ from the latent representation $\mathbf{u}$:
\begin{align}
    L_{\text{adv}}(\mathbf{x}^s) = - \mathbb{E}_{q_\phi(\mathbf{u}|\mathbf{x}^s)}[\log r_\psi(s | \mathbf{u})].
\end{align}
We negate the cross-entropy loss to train the sample generator to prevent the discriminator $r_\psi$ from predicting sensitive attributes. 

We further introduce the property of \textbf{cyclic consistency} to enhance training stability and improve generation quality. Our loss ensures that the counterfactual-counterfactual sample (i.e., double-flipped sample created by our generator) is similar to the original item.
We produce a counterfactual sample $\mathbf{x}_{\text{cnt}}^{s'}$ of the input instance $\mathbf{x}^s$, repeating the three steps below: \looseness=-1
\begin{enumerate}[leftmargin=*]
\item Compute a latent representation $\mathbf{u}$ from the encoder $q_\phi$.
\item Choose the target sensitive attribute $s' (\neq s)$.
\item Reconstruct the counterfactual sample from the decoder $p_\theta$ given the latent variable $\mathbf{u}$ and the target sensitive attribute $s'$. 
\end{enumerate}
The reconstructed sample $\mathbf{x}_{\text{cnt}}^{s'}$ goes back to the generator to produce a counterfactual-counterfactual (or double-flipped) sample $\mathbf{x}_{\text{cyc}}^s$ that shares the identical sensitive attribute with the original input. Our cyclic consistency loss in C-VAE maximizes the likelihood between two instances; $\mathbf{x}^s$, $\mathbf{x}_{\text{cyc}}^s$ (Eq.~\ref{eq:cyc_loss}).
\begin{align}
    convert(\mathbf{x}^{s'}_{\text{cnt}} | \mathbf{x}^s, s') = p_\theta (\mathbf{x}^{s'}_{\text{cnt}} | q_\phi (\mathbf{u} | \mathbf{x}^s), s'), \nonumber \\
    L_{\text{cyc}}(\mathbf{x}^s) = -\mathbb{E}_{convert(\mathbf{x}^{s'}_{\text{cnt}} | \mathbf{x}^s, s')}[\log convert(\mathbf{x}^{s} | \mathbf{x}^{s'}_{\text{cnt}}, s)] \label{eq:cyc_loss}
\end{align}
The total loss to train the sample generator is as follows:
\begin{align}
    L_{\text{c-vae}}(\mathbf{x}^s) = L_{\text{vae}}(\mathbf{x}^s) - L_{\text{adv}}(\mathbf{x}^s) + L_{\text{cyc}}(\mathbf{x}^s).
\end{align}

\subsection{Fairness-aware Contrastive Learning}
The next objective, given the learned sample generator, is to train an encoder that satisfies both group and counterfactual fairness. We use contrastive learning for this task. Let us denote an input query as $\mathbf{x}\in\mathbb{R}^{d}$, and a positive sample and a set of negative samples for contrastive learning as $\mathbf{x}_+$ and $\mathcal{X}_-$, respectively. Typically, positive samples and negative samples are determined by predefined rules. For example, a rule can set a single data instance and its augmented versions as positive, whereas set all others as negative (see description in $\S$\ref{ref:2}). In this work, we define InfoNCE loss in contrastive learning to train the model $F$ (i.e., $F := h_1 \circ g \circ f$ in Figure~\ref{fig:main_training}) as follows: \looseness=-1
\begin{align}
L_{c}(\mathbf{x}, \mathbf{x}_{+}, \mathcal{X}_{-}) &=  -\log {{\text{exp}({\text{sim}(F(\mathbf{x}), F(\mathbf{x}_+)) / \tau}})  \over {\sum_{\mathbf{x}' \in \{\mathbf{x}_{+}\} \cup \mathcal{X}_{-}} \text{exp}(\text{sim}(F(\mathbf{x}), F(\mathbf{x}')) / \tau)}} \label{eq:contrastive_loss} \\
&= - \text{sim}(F(\mathbf{x}), F(\mathbf{x}_+)) / \tau \nonumber \\
&\ \ \ \  + \log{\sum_{\mathbf{x}' \in \{\mathbf{x}_{+}\} \cup \mathcal{X}_{-}} \text{exp}(\text{sim}(F(\mathbf{x}), F(\mathbf{x}')) / \tau)} \\
&= -L_{\text{align}}(\mathbf{x}, \mathbf{x}_+) + L_{\text{distribution}}(\mathbf{x}, \mathbf{x}_{+}, \mathcal{X}_{-}),
\label{eq:simclr_decompose}
\end{align}
where sim($\cdot$) is the function to measure two embeddings' similarity, and $\tau$ is the temperature parameter.

This contrastive loss in Eq.~\ref{eq:simclr_decompose} comprises two terms. First, the alignment loss ($L_{\text{align}}$) encourages the embedding positions of positive pairs to be placed closer. Second, in contrast, the distribution loss ($L_{\text{distribution}}$) matches all instances' embeddings into the prior distribution with a high entropy value. Our model uses the generalized contrastive objective proposed in~\cite{wang2020understanding}, which supports the diverse choice of prior distributions by introducing the optimal transport theory~\cite{bonneel2015sliced} and by changing the distribution loss with Sliced Wasserstein Distance (SWD)~\cite{kolouri2019generalized}. Given the prior distribution over the embedding space $\mathcal{Z}_{prior}$ and the set of embeddings $\mathcal{\tilde{Z}}=\{ F(\mathbf{x}) | \mathbf{x} \in \mathcal{X}_{-}\}$, the loss is formulated as the following equation: \looseness=-1
\begin{align}
L_{\text{gen-}c}(\mathbf{x}, \mathbf{x}_{+}, \mathcal{X}_{-}) = -L_{\text{align}}(\mathbf{x}, \mathbf{x}_{+}) + \text{SWD}(\mathcal{\tilde{Z}}, \mathcal{Z}_{prior}). \label{eq:gen_infonce}
\end{align}

We modify the contrastive loss to jointly meet two fairness criteria. Assume that all instances in $\mathcal{X}_{-}$ are sampled to have the same sensitive attribute $s$ (e.g., gender $=$ female). Then minimizing the distribution loss on $\mathcal{X}_{-}$ (i.e., $\text{SWD}(\mathcal{\tilde{Z}}, \mathcal{Z}_{prior}$) in Eq.~\ref{eq:gen_infonce}) will cause the sensitive group $s$ in embeddings to match the predefined prior distribution $\mathcal{Z}_{prior}$. By iterating this process for every sensitive group (e.g., female and male), our model can produce embeddings of groups that follow the same distribution, $\mathcal{Z}_{prior}$. As a result, embeddings 
become no longer distinguishable by sensitive attributes. Given a batched set of instances $\mathcal{B}_s$ and an input $\mathbf{x}^s$, negative samples are defined as follows: \looseness=-1
\begin{align}
    \mathcal{X}^s_{-} = \{ \mathbf{x} | \mathbf{x} \in \mathcal{B}_s\setminus\{\mathbf{x}^s\} \}.
\end{align}

To ensure counterfactual fairness, our model also considers a counterfactual version of the sample as positive in contrastive learning. Given an input sample $\mathbf{x}^s$, we flip the sensitive attribute $s\rightarrow s'$ (e.g., female to male) and generate the counterfactual sample $\mathbf{x}^{s'}_{\text{cnt}}$ from our C-VAE based sample generator. The alignment loss in Eq.~\ref{eq:gen_infonce} then minimizes the embedding discrepancy between the original and counterfactual data instances. From the positive sample and the set of negative samples ($\mathbf{x}^{s'}_{\text{cnt}}$, $\mathcal{X}^s_{-}$), the fairness-aware contrastive loss for the given input $\mathbf{x}^s$ is defined as follows:
\begin{align}
    L_{\text{fair-cl}}(\mathbf{x}^s) =  L_{\text{gen-}c}(\mathbf{x}^s, \mathbf{x}^{s'}_{\text{cnt}}, \mathcal{X}^s_{-}). \label{eq:contrast_loss}
\end{align}
We train the embedding on top of the Euclidean space with Gaussian prior $\mathcal{Z}_{prior}$. This is different from other contrastive approaches where embeddings are learned over the L2-normalized space~\cite{chen2020simple,he2020momentum}. L2-normalized space used in those approaches does not account for the norm of embeddings during training, which can also be an important clue for sensitive attributes in downstream tasks. However, our choice of the Euclidean space regularizes the norm distribution and removes its dependency on sensitive attributes. The alignment loss $L_{\text{align}}$ is then defined with a negative Euclidean distance as a similarity measure between the original and positive instance. \looseness=-1

\subsection{Self-knowledge Distillation}
The final objective is maintaining the representation quality. We design the self-knowledge distillation loss to reduce the embedding discrepancy between the original and perturbed instances. Inspired by the original literature~\cite{grill2020bootstrap}, we present a Siamese network with two different heads, where each head becomes the student and teacher branches. Then, the model is trained through a prediction task so that the original instance from the teacher branch is highly predictive of the perturbed instances from the student branch. This process is called \textit{self-knowledge distillation}, since the knowledge extracted from the teacher branch is progressively transferred back to the student branch~\cite{kim2021self,tejankar2021isd}. To prevent the model from collapsing into a naive solution (e.g., representation becomes constant for every instance), we let the architecture of student and teacher be asymmetric, restricting the gradient flows of the teacher branch~\cite{chen2021exploring,grill2020bootstrap} (see the bottom part of Figure~\ref{fig:main_training}). Let us denote $f$, $g$, and $h_2$ as the backbone network, the projection head, and the prediction head, respectively. The self-knowledge distillation loss is defined as follows:
\begin{align}
\mathbf{p}_{\text{student}} &= h_2 \circ g \circ f(\mathbf{x}^s_{\text{pert}}), \ \ \ \ \ \mathbf{z}_{\text{teacher}} = sg \circ g \circ f(\mathbf{x}^s)  \nonumber \\
&L_{\text{self-kd}}(\mathbf{x}^s) = - {\mathbf{p}_{\text{student}} \over || \mathbf{p}_{\text{student}} ||_2} \cdot {\mathbf{z}_{\text{teacher}}  \over || \mathbf{z}_{\text{teacher}} ||_2}, \label{eq:self-kd}
\end{align}
where $\mathbf{x}^s_{\text{pert}}$ is a perturbed version of original instance $\mathbf{x}^s$ and $sg$ represents the stop-gradient operation.

When perturbing the instance, we propose \tabmix{} for our augmentation strategy. Given an input sample $\mathbf{x}^s$, the generator randomly masks $k$ features and replaces their values from other instances, as shown in Figure~\ref{fig:tabmix}.\footnote{We mask 50\% of the non-sensitive attributes. The masking ratio is not critical to the downstream task.} Let $\odot$ denote an element-wise multiplicator and $\mathbf{m} \in \{0,1\}^{d}$ a binary mask vector indicating which feature to replace. The mixing operation is defined as:
\begin{align}
    \text{TabMix}(\mathbf{x}_i, \mathbf{x}_j) = \mathbf{m} \odot \mathbf{x}_i + (1 - \mathbf{m}) \odot \mathbf{x}_j. \label{eq:tabmix}
\end{align}
\tabmix{} does not need to consider the scale of numeric variables; hence it is easily applicable to various datasets. The perturbed instance $\mathbf{x}^s_{\text{pert}}$ is obtained as follows: 
\begin{align}
    \mathbf{x'} &\sim \text{Sample}(\mathcal{D} \setminus \mathbf{x}^s) \\
    \mathbf{x}^s_{\text{pert}} &= \text{TabMix}(\mathbf{x}^s, \mathbf{x'}).
\end{align}

Finally, the loss function of the entire process is the sum of two losses: fairness-aware contrastive loss ($L_{\text{fair-cl}}$) and self-knowledge distillation loss ($L_{\text{self-kd}}$) as in Eq.~\ref{eq:final_loss}.
\begin{align}
    L_{\text{total}} = {1 \over |\mathcal{D}|} \sum_{\mathbf{x}^s \in \mathcal{D}}(L_{\text{fair-cl}}(\mathbf{x}^s) + L_{\text{self-kd}}(\mathbf{x}^s)) \label{eq:final_loss}
\end{align}

\section{Experiment}

We compared \model{} with the latest models on multiple fairness-aware tabular datasets. Component analyses were performed to test the effect of each module. We also present qualitative analyses and case studies on how the model handles sensitive attributes. 

\subsection{Performance Evaluation}
\textbf{Datasets:} We use the following datasets: (1) UCI Adult~\cite{asuncion2007uci} contains 48,842 samples along with label information that indicates whether a given individual makes over 50K per year as a downstream task; (2) UCI German Credit~\cite{asuncion2007uci} includes 1,000 samples with 20 attributes and aims to predict credit approvals; (3) COMPAS~\cite{compas2016} includes 6,172 samples and is used to predict recidivism risk (i.e., the risk of a criminal defendant committing a crime again) of given individuals; (4) LSAC~\cite{wightman1998lsac} contains 22,407 samples and estimates whether a given individual will pass the law school admission; (5) Students Performance (Students) ~\cite{cortez2008using} consists of 649 samples and predicts the grade in exam; (6) Communities~\cite{redmond2002data} consists of 1,994 samples and estimates the number of violent crimes per 100K population.

We split each dataset into disjoint training and test sets. Embedding learning and counterfactual sample generation are based on the knowledge of the training set; evaluations are based on the test set. Table~\ref{tab:data_description} summarizes statistics of each dataset. \smallskip

\begin{table}[t!]
\setlength{\tabcolsep}{2.5pt}
\centering
\frenchspacing
\caption{Data descriptions of six datasets}
\resizebox{1.00\linewidth}{!}{%
\begin{tabular}{lrrlcl}
\toprule
\multicolumn{1}{c}{Dataset} &
\multicolumn{1}{c}{$\#$ samples} &
\multicolumn{1}{c}{$\#$ attr.} & 
\multicolumn{1}{c}{Sensitive attr.} & \multicolumn{1}{c}{Split} & \multicolumn{1}{c}{Task} \\ \midrule
Adult           & 48,842   & 14      & gender, race    & 2:1   & classification             \\
Credit   & 1,000    & 20      & gender          & 4:1  & classification             \\
COMPAS              & 6,172    & 7       & gender, race    & 4:1 & classification             \\
LSAC                & 22,407   & 12      & gender, race    & 4:1   & classification           \\
Students & 649 & 33 & gender & 4:1 & regression \\
Communities & 1,994 & 128 & race & 4:1 & regression \\
\bottomrule
\end{tabular}
\label{tab:data_description}
}
\end{table}

\noindent
\textbf{Evaluation: } The evaluation uses embeddings learned from the last ten epochs, and the averaged results of five metrics below:
\begin{itemize}[leftmargin=5mm]
    \item \textbf{\textsf{AUC}}, the area under the receiver operating characteristics, measures the prediction performance of the downstream classification task (i.e., the performance of the binary classifier to distinguish between cases and non-cases). If this value is 1, the model distinguishes the target variable from input instances with absolute precision.

    \item \textbf{\textsf{RMSE}}, the root mean squared error, measures the deviance between the prediction and ground-truth for the regression task. \looseness=-1
    
    \item \textbf{\textsf{Demographic Parity Distance ($\Delta DP$)}} is a group fairness metric and is defined as the expected absolute difference between the predictions of protected groups. Given a set of sensitive attributes $\mathcal{S}$, the definition of $\Delta DP$ is as follows ($s\neq s'$):
    \begin{align}
        \Delta DP = \mathbb{E}_{s,s' \in S}[|P(\hat{Y}=1 | s) - P(\hat{Y}=1 | s')|].
    \end{align}
    
    \item \textbf{\textsf{Equalized Odds ($\Delta EO$)}}, also known as equality of opportunity, is a group fairness metric based on the expected difference between the estimated positive rates for two protected groups. Given the set of sensitive attributes $\mathcal{S}$, $\Delta EO$ is defined as follows ($y \in \{0, 1\}$, $s\neq s'$):
    \begin{align}
        \Delta EO = \mathbb{E}_{s,s' \in S}[|P(\hat{Y}=1 | s, Y=y) - P(\hat{Y}=1 | &s', Y=y)|].
    \end{align}
    
    \item \textbf{\textsf{Counterfactual Parity Distance ($\Delta CP$)}} is a metric for counterfactual fairness and measures the prediction parity between two instances in counterfactual pair relationships. Assume that $(\mathbf{x}, \hat{\mathbf{x}})$ are from a set of counterfactual pairs $\mathcal{D} \times \hat{\mathcal{D}}$. Then $\Delta CP$ is defined as follows:
    \begin{align}
        \Delta CP = \mathbb{E}_{(\mathbf{x}, \hat{\mathbf{x}}) \in \mathcal{D} \times \hat{\mathcal{D}}}[|P(\hat{Y}=1 | \mathbf{x}) - P(\hat{Y}=1 | \hat{\mathbf{x}}) |].
    \end{align}
\end{itemize}
\noindent
\textbf{Baselines: } A total of nine baselines are employed. The first two directly utilize raw datasets to classify target variables, given that the downstream task is already known. They differ by input structure: (1) original data (LR) and (2) original data concatenated with the synthetic counterfactual samples (C-LR). The next one is (3) SCARF~\cite{bahri2021scarf}, an unsupervised contrastive learning method that does not consider any fairness requirement during training. The remaining six learn fair representations from data without any information on downstream tasks. They utilize different unsupervised representation learning techniques with fairness-aware objectives. (4) VFAE~\cite{louizos2016variational} introduces the maximum mean discrepancy term on top of the variational autoencoder to produce fair representations. (5) LAFTR~\cite{madras2018learning} adopts an adversarial approach to avoid unfair predictions from embeddings. (6,7) MIFR and L-MIFR~\cite{song2019learning} learn the controllable fair representation through mutual information. The two differ in the use of the Lagrangian dual optimization method. (8,9) C-InfoNCE and WeaC-InfoNCE~\cite{tsai2021conditional} maximize the conditional mutual information within representations. The two differ in how they introduce the sensitive variable in the InfoNCE objective. For all methods, we use a logistic regression model for the downstream classification task and a Random Forest regression model for the downstream regression task as a base predictor. The predictor is learned on top of either raw dataset (1--2) or learned embeddings (3--9). \looseness=-1 \smallskip

\noindent
\textbf{Implementation details: } We trained the multi-layer perceptron (MLP) model. A three-layer MLP was used for the backbone network $f$, and a two-layer MLP for both projection and prediction heads $g$, ($h_1$, $h_2$). ReLU activation function was used for all architectures, with the training of 200 epochs and a batch size of 128. Adam optimizer with a learning rate of 1e-3 and a weight decay factor of 1e-6 was utilized. For the counterfactual sample generator, three-layer, two-layer, and two-layer MLPs were utilized for the encoder, decoder, and discriminator, respectively. The sample generator was trained for 600 epochs with the Adam optimizer. Mode-specific normalization~\cite{xu2019modeling} was used for preprocessing categorical variables. \looseness=-1 \smallskip

\begin{table}[h!]
\caption{Performance comparison summaries among fairness-aware baselines and \model{}. Averaged rank for each evaluation metric across six datasets is reported. 
}
\centering
\label{table:summary}
\resizebox{0.92\columnwidth}{!}{
\begin{tabular}{lccccc}
\toprule
Method & AUC/RMSE & $\Delta DP$ & $\Delta EO$ & $\Delta CP$ & Total \\ \midrule
VFAE & 3.2 & 5.8 & 5.0 & 4.2 & 4.6 \\
LAFTR & \textbf{2.7} & 6.2 & 5.5 & 2.8 & 4.3 \\
MIFR & 6.0 & 3.0 & 3.8 & 4.6 & 4.4 \\
L-MIFR & 5.3 & 3.2 & 3.3 & 3.8 & 3.9 \\
C-InfoNCE & 3.2 & 4.5 & 3.3 & 5.2 & 4.1 \\
WeaC-InfoNCE & 4.7 & 2.8 & 4.3 & 4.8 & 4.2 \\ \midrule
\textbf{\model{}} & 3.0 & \textbf{2.5} & \textbf{2.5} & \textbf{2.2} & \textbf{2.6} \\ \bottomrule
\end{tabular}
}
\end{table}
\begin{table*}[t!]
\setlength{\tabcolsep}{2.5pt}
\caption{Detailed results over downstream classification tasks. Across the four datasets, \textit{gender} is considered a sensitive attribute to protect. Up-arrow ($\uparrow$) denotes that a higher value is better, and down-arrow ($\downarrow$) is vice versa. The best results among fair representation learning baselines are highlighted in bold.
}
\centering
\label{table:classificationresults}
\resizebox{2.1\columnwidth}{!}{
\begin{tabular}{@{}lcccc|cccc|cccc|cccc@{}}
\toprule
\multicolumn{1}{l}{\multirow{2}{*}{Method}} & \multicolumn{4}{c|}{UCI Adult} & \multicolumn{4}{c|}{UCI German credit} & \multicolumn{4}{c|}{COMPAS} & \multicolumn{4}{c}{LSAC} \\ \cmidrule(l){2-17} 
\multicolumn{1}{c}{} &  \multicolumn{1}{c}{AUC ($\uparrow$)} & \multicolumn{1}{c}{$\Delta DP$ ($\downarrow$)} & \multicolumn{1}{c}{$\Delta EO$ ($\downarrow$)} & \multicolumn{1}{c|}{$\Delta CP$ ($\downarrow$)} & \multicolumn{1}{c}{AUC} & \multicolumn{1}{c}{$\Delta DP$} & \multicolumn{1}{c}{$\Delta EO$} & \multicolumn{1}{c|}{$\Delta CP$} & \multicolumn{1}{c}{AUC} & \multicolumn{1}{c}{$\Delta DP$} & \multicolumn{1}{c}{$\Delta EO$} & \multicolumn{1}{c|}{$\Delta CP$} & \multicolumn{1}{c}{AUC} & \multicolumn{1}{c}{$\Delta DP$} & \multicolumn{1}{c}{$\Delta EO$} & \multicolumn{1}{c}{$\Delta CP$} \\ \midrule
LR & 0.9055 & 0.1931 & 0.1763 & 0.0812 & 0.7580 & 0.0653 & 0.1066 & 0.1765 & 0.7354 & 0.1175 & 0.1733 & 0.0873 & 0.8515 & 0.0463 & 0.0378 & 0.0493   \\
C-LR &  0.9023 & 0.1610 & 0.1679 & 0.0568 & 0.7252 & 0.0161 & 0.0879 & 0.1416 & 0.7321 & 0.0441 & 0.0592 & 0.0286 & 0.8513 & 0.0621 & 0.0361 & 0.0327 \\ 
SCARF & 0.9012 & 0.1905 & 0.1929 & 0.1097 & 0.7459 & 0.0719 & 0.1443 & 0.1460 & 0.7384 & 0.1155 & 0.1747 & 0.1029 & 0.8544 & 0.0716 & 0.0357 & 0.0477 \\ \midrule
VFAE & 0.8249 & 0.1764 & 0.1995 & 0.1126 & 0.6582 & 0.0365 & 0.0514 & 0.2785 & 0.7089 & 0.0286 & 0.1069 & 0.1226 & \textbf{0.8275} & 0.0725 & 0.0293 & 0.0569  \\
LAFTR & \textbf{0.8943} & 0.1727 & 0.1720 & 0.0746 & 0.6845 & 0.0223 & 0.0282 & 0.1917 & 0.7142 & 0.0486 & 0.1055 & 0.0562 & 0.6380 & 0.0731 & 0.0172 & 0.0428 \\
MIFR & 0.8162 & 0.0427 & 0.0649 & 0.1238 & 0.5463 & 0.0113 & 0.0297 & 0.1241 & 0.6112 & \textbf{0.0081} & 0.0423 & 0.1049 & 0.7672 & 0.0676 & \textbf{0.0142} & 0.0618 \\
L-MIFR & 0.8211 & 0.0337 & 0.0578 & 0.1408 & 0.5703 & \textbf{0.0045} & 0.0143 & 0.1217 & 0.6243 & 0.0100 & \textbf{0.0409} & 0.0996 & 0.7694 & 0.0680 & 0.0152 & 0.0613 \\
C-InfoNCE & 0.8252 & 0.0884 & 0.0909 & 0.1492 & 0.6719 & 0.0206 & 0.0191 & 0.1141 & 0.7192 & 0.0495 & 0.0859 & 0.0793 & 0.7977 & 0.0743 & 0.0149 & 0.2647  \\
WeaC-InfoNCE & 0.8153 & 0.0786 & 0.0802 & 0.1895 & 0.6872 & 0.0164 & 0.0165 & \textbf{0.1023} & 0.7130 & 0.0394 & 0.0871 & 0.0750 & 0.7914 & 0.0648 & 0.0412 & 0.2048  \\ \midrule
\textbf{\model{}} & 0.8002 & \textbf{0.0285} & \textbf{0.0435} & \textbf{0.0644} & \textbf{0.7292} & 0.0061 & \textbf{0.0130} & 0.1454 & \textbf{0.7235} & 0.0401 & 0.0622 & \textbf{0.0409} & 0.7790 & \textbf{0.0527} & 0.0239 & \textbf{0.0361} \\ \bottomrule
\end{tabular}
}
\end{table*}
\begin{table}[t!]
\setlength{\tabcolsep}{2.5pt}
\caption{Detailed results over downstream regression tasks. The best results among fair representation learning baselines are highlighted in bold.
}
\centering
\label{table:regressionresults}
\resizebox{1\columnwidth}{!}{
\begin{tabular}{@{}lccc|ccc@{}}
\toprule
\multicolumn{1}{l}{\multirow{2}{*}{Method}} & \multicolumn{3}{c|}{Students} & \multicolumn{3}{c}{Communities} \\ \cmidrule(l){2-7} 
\multicolumn{1}{c}{} &  \multicolumn{1}{c}{RMSE ($\downarrow$)} & \multicolumn{1}{c}{$\Delta DP$ ($\downarrow$)} & \multicolumn{1}{c|}{$\Delta CP$ ($\downarrow$)} &  \multicolumn{1}{c}{RMSE} & \multicolumn{1}{c}{$\Delta DP$} & \multicolumn{1}{c}{$\Delta CP$} \\ \midrule
LR & 0.0392 & 0.0177 & 0.0794 & 0.0145 & 0.2146 & 0.1559 \\
C-LR & 0.0399 & 0.0081 & 0.0777 & 0.0141 & 0.1643 & 0.0854 \\
SCARF & 0.0746 & 0.0196 & 0.0412 & 0.0821 & 0.1631 & 0.0881 \\ \midrule
VFAE & 0.0425 & 0.0084 & \textbf{0.0519} & \textbf{0.0191} & 0.1714 & 0.0844 \\
LAFTR & 0.0371 & 0.0223 & 0.0758 & 0.0194 & 0.1725 & 0.1311 \\
MIFR & 0.0475 & 0.0056 & 0.0785 & 0.0261 & 0.1139 & 0.0916 \\
L-MIFR & 0.0485 & 0.0063 & 0.0797 & 0.0258 & 0.1161 & 0.0902 \\
C-InfoNCE & 0.0415 & \textbf{0.0049} & 0.0626 & 0.0317 & 0.0841 & 0.0884 \\
WeaC-InfoNCE & 0.0407 & 0.0054 & 0.0676 & 0.0319 & \textbf{0.0813} & 0.0891 \\ \midrule
\textbf{DualFair} & \textbf{0.0382} & 0.0054 & 0.0735 & 0.0247 & 0.1130 & \textbf{0.0816} \\ \bottomrule
\end{tabular}
}
\end{table}

\noindent
\textbf{Results: } 
\model{} outperforms other baselines in terms of the averaged rank for each evaluation metric. Table~\ref{table:summary} shows that our model's embedding maintains its prediction performance (i.e., AUC and RMSE) while successfully removing any bias related to sensitive information from the embedding. \looseness=-1

Tables~\ref{table:classificationresults} and \ref{table:regressionresults} report the detailed performance of \model{} and other baselines on downstream classification and regression tasks. According to the results, naively adding the counterfactual samples (i.e., C-LR) is insufficient to handle bias from sensitive information. Similarly, the fair embedding learning baselines such as VFAE, L-MIFR, or WeaC-InfoNCE fail to remove sensitive information entirely nor generate a fair embedding without losing critical information for the downstream task. These findings suggest that \model{} achieves both the performance and fairness requirements in a single training.

\subsection{Component Analyses}
We performed an ablation study by repeatedly assessing and comparing the models after removing each component. We also examined the counterfactual sample quality.  \smallskip

\noindent
\textbf{Ablation study: } \model{} utilizes two learning objectives: fairness-aware contrastive loss to ensure both group and counterfactual fairness, and self-knowledge distillation loss to ensure the representation quality. Ablations remove each loss objective from the full model to assess the unique contribution. Table~\ref{Tab:Ablation} reports the results based on the UCI Adult dataset. The full model achieves the best balance between fairness and prediction performance, implying that each loss plays a unique role in designing fair representations. The result without self-knowledge distillation shows that our fairness-aware contrastive loss effectively improves fairness while there is a trade-off for prediction performance. On the other hand, the experiment only with self-knowledge distillation loss (i.e., w/o $L_{\text{align}}$ and $L_{\text{distribution}}$) produces opposite result. The result without only alignment loss, which is in charge of counterfactual fairness, contrasts the role of two losses in fairness-aware contrastive objective; Counterfactual fairness does not have any improvement while group fairness has been reduced. 
Additionally, we experimented using Gaussian noise and dropout as alternative augmentation strategies for TabMix. This ablation reduced the prediction accuracy (AUC: 0.80$\rightarrow$0.78, 0.75) while maintaining fairness (DP: 0.03$\rightarrow$0.03, 0.02). \smallskip

\begin{table}[!t]
\centering
\caption{Ablation results of \model{} on UCI Adult dataset. $L_{\text{align}}$ and $L_{\text{distribution}}$ are from fairness-aware contrastive loss while $L_{\text{self-kd}}$ is self-knowledge distillation loss.}
\begin{tabular}{lcccc}
\toprule
Setup & AUC & $\Delta DP$ & $\Delta EO$ & $\Delta CP$ \\ \midrule
\model{}  & 0.80 &  0.03 & 0.04 & 0.06 \\
w/o $L_{\text{self-kd}}$  & 0.77 &  0.02 & 0.03 & 0.05 \\
w/o $L_{\text{align}}$             & 0.78 & 0.04 & 0.04 & 0.08 \\ 
w/o $L_{\text{align}}$ and $L_{\text{distribution}}$ & 0.81 & 0.12 & 0.12 & 0.08\\
\bottomrule 
\end{tabular}
\label{Tab:Ablation}
\end{table}
\begin{table}[!t]
\setlength{\tabcolsep}{2.5pt}
\centering
\caption{The model trained with counterfactual samples shows good classification performance on the original data, proving that counterfactual samples are augmented well.}
\resizebox{\linewidth}{!}{
\begin{tabular}{lcccccccc}
\toprule
 & \multicolumn{2}{c}{Adult} & \multicolumn{2}{c}{Credit} & \multicolumn{2}{c}{Compas} & \multicolumn{2}{c}{LSAC} \\ \cmidrule{2-9}
Training set  & AUC & F1 & AUC & F1 & AUC & F1 & AUC & F1 \\ \midrule
Original          & 0.91 & 0.78 & 0.76 & 0.68 & 0.74 & 0.68 & 0.85 & 0.65 \\ 
Counterfactual & 0.89 & 0.76 & 0.75 & 0.62 & 0.73 & 0.67 & 0.84 & 0.63 \\
\bottomrule 
\end{tabular}}
\label{Tab:counterfactual-only}
\end{table}

\noindent
\textbf{Counterfactual samples: }
To test the quality of the generated counterfactual samples, we examined if the target variable is predictable, even if the model is trained only with the synthetic counterfactual samples. Table~\ref{Tab:counterfactual-only} reports the logistic regression performance that changes the training set from the original data. The latter model is on par with the model trained with the original dataset. \looseness=-1

\begin{figure}[t!]
\centering
\begin{subfigure}[t]{0.21\textwidth}
       \centering\includegraphics[height=4.51cm]{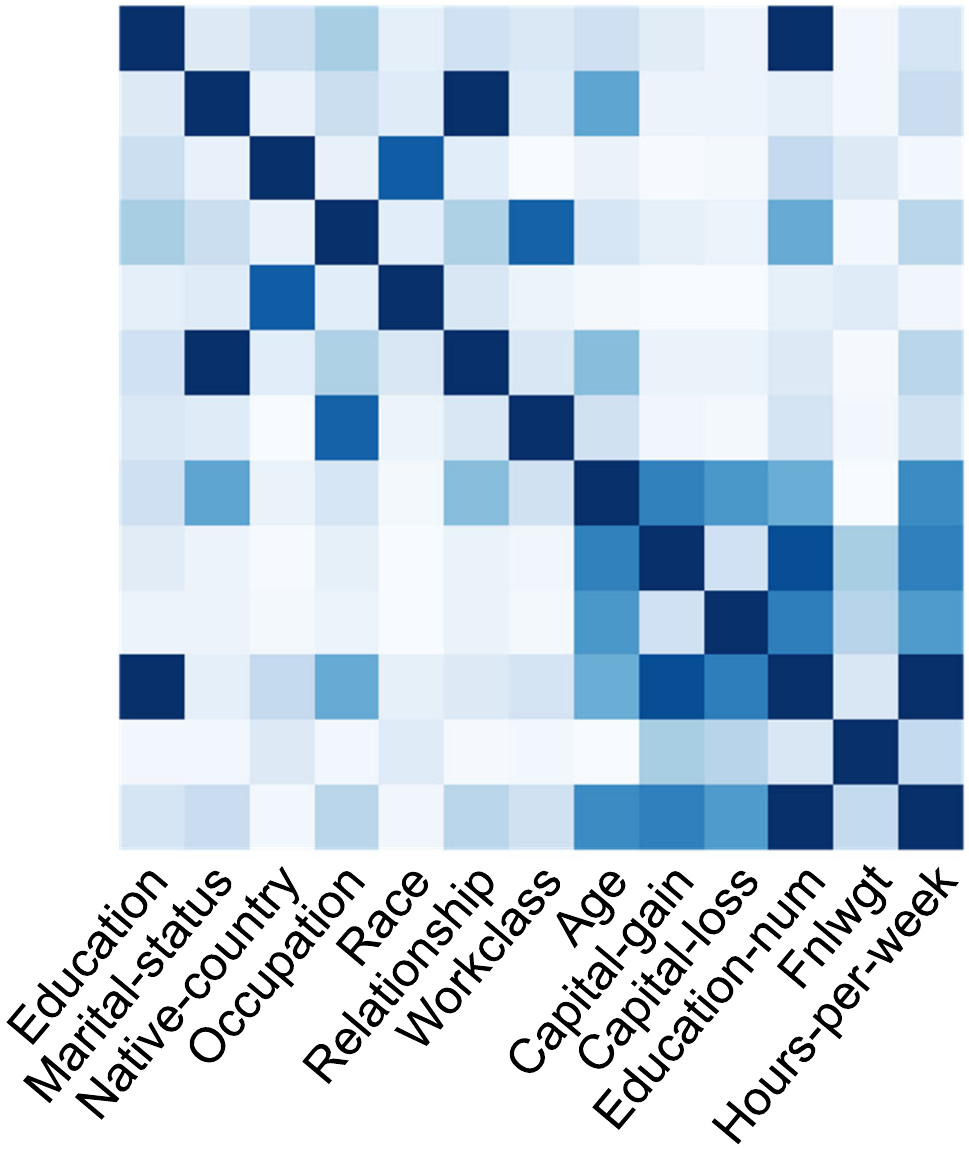}
      \caption{Original UCI Adult}
      \label{fig:corr_original}
\end{subfigure}
\hspace{1mm}
\begin{subfigure}[t]{0.25\textwidth}
       \centering\includegraphics[height=4.5cm]{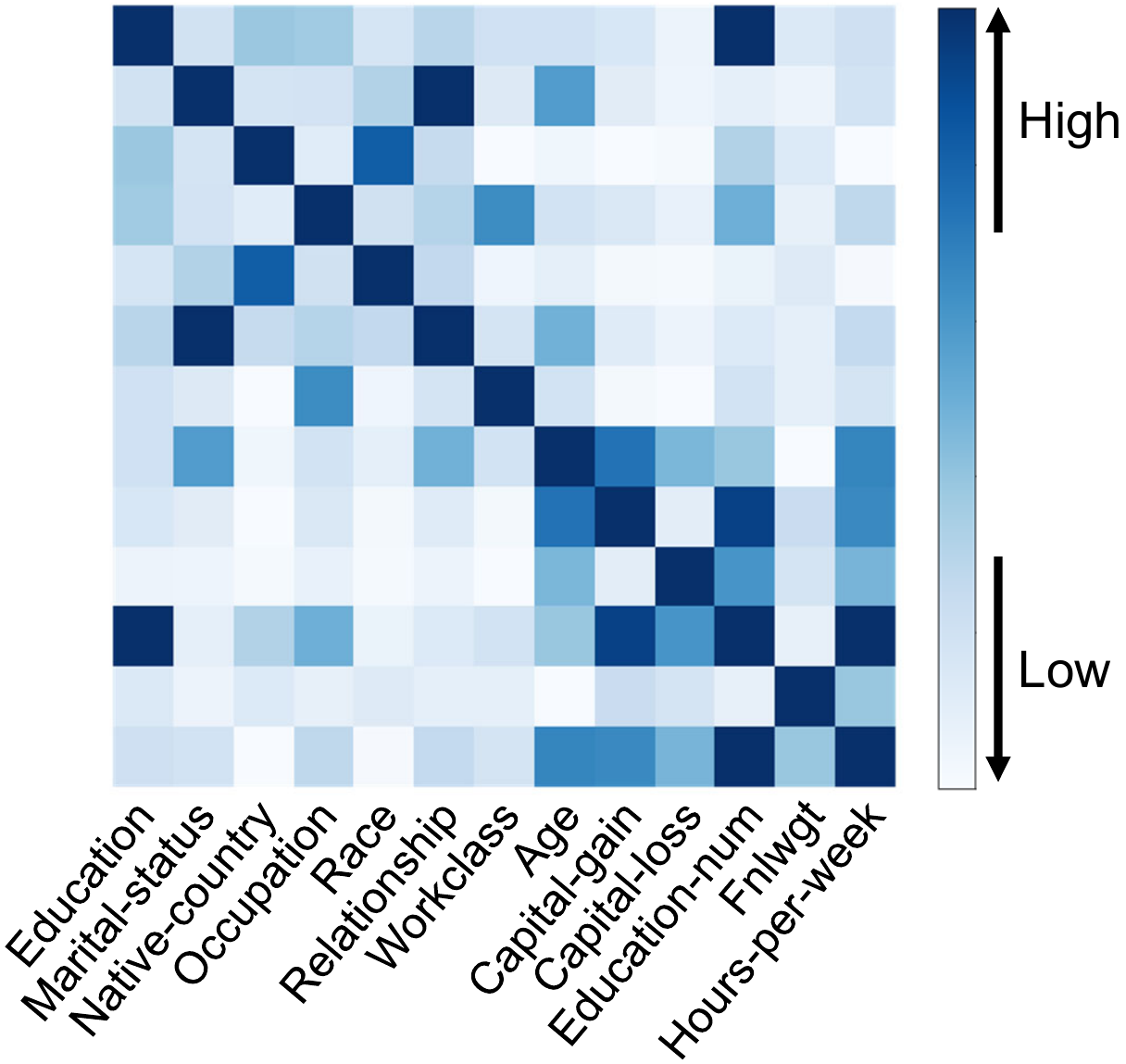}
      \caption{Counterfactual synthetic data}
      \label{fig:corr_counter}
\end{subfigure}
\caption{Correlation heatmaps show that (a) original UCI adult dataset and (b) synthetic dataset from our counterfactual sample generator---follow similar patterns.}
\label{fig:corr_all}
\end{figure}

We next examined if feature correlations are maintained for counterfactual samples. Figure~\ref{fig:corr_all} shows the correlation matrix between features in the original UCI adult dataset and the counterfactual dataset. Pearson correlation is used between continuous variables, and Cramer's V value is utilized between categorical variables. To measure the correlation between categorical and continuous variables, we label-encode the categorical variables to their continuous counterparts and compute the Pearson correlation with original continuous values. Two matrices show a remarkable resemblance, indicating that the relationship between features in the counterfactual samples is well-maintained.

\begin{figure}[t!]
\centering
\begin{subfigure}[t]{0.23\textwidth}
\captionsetup{justification=centering}
       \centering\includegraphics[height=3.2cm]{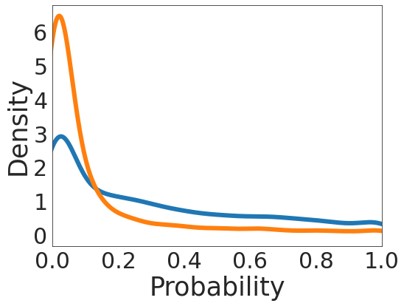}
      \caption{Standard ($\Delta DP$=0.19)}
      \label{fig:density_scarf}
\end{subfigure}
\begin{subfigure}[t]{0.23\textwidth}
\captionsetup{justification=centering}
       \centering\includegraphics[height=3.2cm]{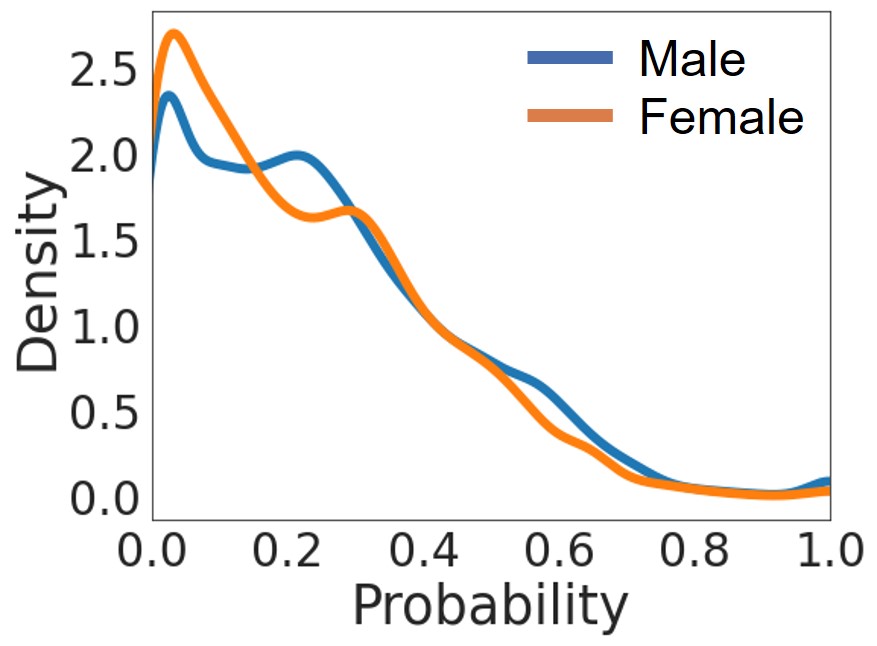}
      \caption{\model{} ($\Delta DP$=0.02)}
      \label{fig:density_ours}
\end{subfigure}
\caption{Income prediction results between the male and female groups are similar when we use \model{}'s embeddings. We compared two models, (a) standard logistic regression on top of original data and (b) logistic regression on top of \model{} embedding, on the UCI Adult dataset.
}
\label{fig:density}
\end{figure}

\subsection{Qualitative Analysis}
We visually check the learned representations to investigate how well our model handles sensitive information. Figure~\ref{fig:density} shows how debiasing achieves group fairness in downstream predictions. It compares the prediction results on UCI Adult for two models: standard logistic regression applied on the raw dataset and the same regression model using the \model{} embeddings. After debiasing, there is almost no difference in the distribution of predicted income between male and female groups, as depicted in Figure~\ref{fig:density_ours}. However, two probability density functions in Figure~\ref{fig:density_scarf} appear substantially different in the standard model. These results confirm the outstanding debiasing potential of the proposed model. \looseness=-1

\begin{figure}[t!]
\centering
\begin{subfigure}[t]{0.23\textwidth}
\captionsetup{justification=centering}
       \centering\includegraphics[height=3.2cm]{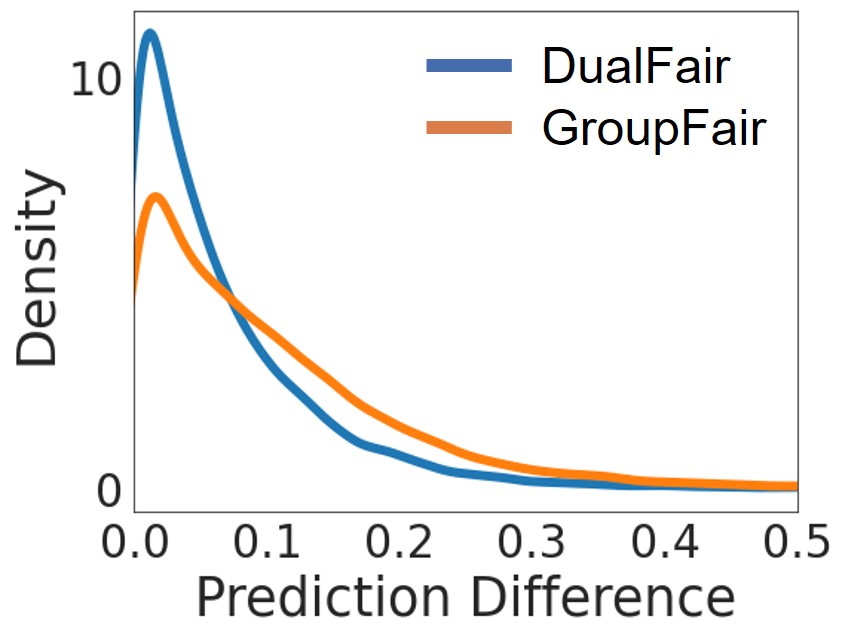}
      \caption{Histogram of $\Delta CP$}
      \label{fig:case_study1}
\end{subfigure}
\begin{subfigure}[t]{0.23\textwidth}
\captionsetup{justification=centering}
       \centering\includegraphics[height=3.2cm]{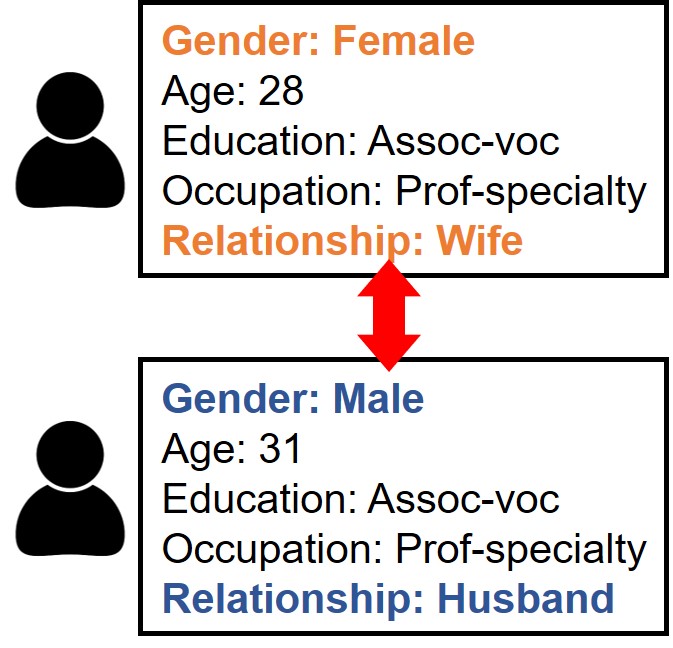}
      \caption{Counterfactual example}
      \label{fig:case_study2}
\end{subfigure}
\caption{Additional findings on the counterfactual model: (a) Comparison of the $\Delta CP$ value between \model{} and its ablation \textsf{GroupFair} which omits the alignment loss between counterfactual pairs. (b) Observable variables affected by sensitive attributes are also changed in the counterfactual example. \looseness=-1}
\label{fig:case_study}
\end{figure}

We also investigate how well \model{} achieves counterfactual fairness by examining the difference in the $\Delta CP$ values of the proposed model and its ablation \textsf{GroupFair}, which omits the alignment loss between counterfactual pairs in fairness-aware contrastive loss (Eq.~\ref{eq:contrast_loss}). Experimental comparison over UCI Adult in Figure~\ref{fig:case_study1} shows a smaller prediction difference for the full model, indicating that the omitted loss is effective in debiasing sensitive attributes at the individual level. One of the counterfactual pairs from the model is illustrated in Figure~\ref{fig:case_study2}. Note that in generating counterfactual examples, we do not simply flip sensitive attributes (i.e., gender) but also observable variables (i.e., age, relationship) change together. When we compute $\Delta CP$ for the example case from two models, we confirm that the proposed \model{} satisfies the fairness concerns to some extent (prediction probability for original: 0.22 vs. counterparts: 0.24). Meanwhile, \textsf{GroupFair} fails to debias gender information and gives a higher score to the male counterparts (0.24 vs. 0.45). \looseness=-1

\section{Conclusion}
We presented \model{}, a self-supervised embedding learning model that de-biases sensitive data attributes without any prior information on downstream tasks. Its design includes a unique fairness-aware contrastive loss and self-knowledge distillation technique.
%
Experiments confirm that \model{} generates rich data representations that ensure both group fairness and counterfactual fairness.
%
Our model is applicable to various Web applications, including classification, ranking, recommendation, and text generation tasks.

The algorithmic bias observed in search engine results and social media platforms has reinforced the need for a clear policy for protecting sensitive attributes. However, the problem is more complex because bias can also exist in other domains, such as in natural language processing (e.g., Q\&A generation, chatbot). The recently released `AI Bill of Rights' blueprint from the White House states that ``you should not face discrimination by algorithms and systems should be used and designed in an equitable way.'' We believe that our self-supervised learning with debiasing techniques can serve as a building block for advancing the performance and fairness requirements of real-world web applications.
\looseness=-1

\section*{Acknowledgement}
This research was supported by the Institute for Basic Science (IBS-R029-C2, IBS-R029-Y4), the Potential Individuals Global Training Program (2022-00155958) by the Ministry of Science and ICT in Korea, and Microsoft Research Asia.
 

\bibliographystyle{ACM-Reference-Format}
\typeout{}
\balance
\bibliography{_z_bibliography}

\begin{table*}[!ht]
\caption{Additional results over downstream classification tasks in which \textit{race} is set as a target-sensitive attribute to protect. Performances are reported with four evaluation metrics (AUROC, $\Delta DP$, $\Delta EO$, and $\Delta CP$) over three datasets.
}
\centering
\resizebox{1.83\columnwidth}{!}{
\begin{tabular}{@{}lcccc|cccc|cccc@{}}
\toprule
\multicolumn{1}{l}{\multirow{2}{*}{Method}} & \multicolumn{4}{c|}{UCI Adult} &\multicolumn{4}{c|}{COMPAS} & \multicolumn{4}{c}{LSAC} \\ \cmidrule(l){2-13} 
\multicolumn{1}{c}{} &  \multicolumn{1}{c}{AUROC ($\uparrow$)} & \multicolumn{1}{c}{$\Delta DP$ ($\downarrow$)} & \multicolumn{1}{c}{$\Delta EO$ ($\downarrow$)} & \multicolumn{1}{c|}{$\Delta CP$ ($\downarrow$)} & \multicolumn{1}{c}{AUROC} & \multicolumn{1}{c}{$\Delta DP$} & \multicolumn{1}{c}{$\Delta EO$} & \multicolumn{1}{c|}{$\Delta CP$} & \multicolumn{1}{c}{AUROC} & \multicolumn{1}{c}{$\Delta DP$} & \multicolumn{1}{c}{$\Delta EO$} & \multicolumn{1}{c}{$\Delta CP$} \\ \midrule
LR  & 0.9055 & 0.0867 & 0.1028 & 0.0771 & 0.7354 & 0.1214 & 0.2515 & 0.0608 & 0.8515 & 0.1306 & 0.1081 & 0.0448 \\
C-LR  & 0.9046 & 0.0729 & 0.0854 & 0.0490 & 0.7353 & 0.1050 & 0.3222 & 0.0273 & 0.8504 & 0.1499 & 0.1462 & 0.0293 \\ 
SCARF  & 0.9011 & 0.0792 & 0.0928 & 0.1058 & 0.7418 & 0.1507 & 0.2949 & 0.0802 & 0.8515 & 0.1399 & 0.1351 & 0.0523 \\ \midrule
VFAE & 0.8445 & 0.0945 & 0.0586 & 0.0746 & 0.7039 & 0.1319 & 0.1664 & 0.3115 & 0.8218 & 0.0595 & 0.1472 & 0.1314 \\
LAFTR & 0.8940 & 0.0652 & 0.0667 & 0.0878 & 0.7175 & 0.0378 & 0.1373 & 0.3257 & 0.7213 & 0.0578 & 0.1031 & 0.0129 \\
MIFR & 0.8297 & 0.0931 & 0.0255 & 0.0463 & 0.6454 & 0.1035 & 0.0099 & 0.2104 & 0.7129 & 0.0579 & 0.0181 & 0.0084 \\
L-MIFR & 0.8392 & 0.1044 & 0.0250 & 0.0455 & 0.6895 & 0.1077 & 0.0593 & 0.1819 & 0.7188 & 0.0566 & 0.0277 & 0.0082 \\
C-InfoNCE & 0.8758 & 0.1854 & 0.0673 & 0.0716 & 0.6256 & 0.0462 & 0.0327 & 0.2091 & 0.6548 & 0.0647 & 0.0141 & 0.0052 \\
WeaC-InfoNCE & 0.8843 & 0.1911 & 0.0687 & 0.0895 & 0.6084 & 0.0325 & 0.0313 & 0.0898 & 0.7087 & 0.1015 & 0.0267 & 0.0118 \\ \midrule
\textbf{\model{}} & 0.8067 & 0.0527 & 0.0205 & 0.0355 & 0.7215 & 0.0376 & 0.1096 & 0.3112 & 0.7553 & 0.0491 & 0.0489 & 0.0451 \\ \bottomrule
\end{tabular}
}
\label{table:overallresults_race}
\end{table*}

\newpage

\newpage

\appendix
\section{Appendix}

\subsection{Implementation Details}
\textbf{\model{}: }
\model{} exploits the multi-layer perceptron architecture for both the main model and counterfactual sample generator. The main model consists of the backbone network ($f$), the projection head ($g$), and two prediction heads ($h_1$, $h_2$). We chose all components to be ReLU networks, whose hidden dimension is set to 256. The backbone network has three layers, and all heads have two layers. \model{} is trained 200 epochs, and its batch size is set to 128. The Adam optimizer with a learning rate of 1e-3 and a weight decay factor of 1e-6 is adopted. We apply one-hot encoding to discrete variables for input preprocessing and z-score scaling to continuous variables. \looseness=-1 \smallskip

\noindent
\textbf{Counterfactual sample generator: }
We introduce the C-VAE model to generate counterfactual samples. When preprocessing the input data, discrete values are encoded into the one-hot vector, and every continuous value is converted to a vector of probability density via mode-specific normalization~\cite{xu2019modeling}. Mode-specific normalization uses a variational Gaussian mixture model (VGM) to estimate the number of modes and fit the Gaussian mixture model on top of the target distribution. Then, the probability density of each mode is computed for the normalization. We discovered that mode-specific normalization improves counterfactual sample quality more than naive min-max normalization. The counterfactual sample generator is trained for 600 epochs, and the batch size is set to 256. The weight between reconstruction loss and probability distribution loss is set to 2:1. \looseness=-1 \smallskip


\noindent
\textbf{Computational complexity: } 
We used four A100 GPUs for all experiments. Our model took 20\% more training time than WeaC-InfoNCE (11 vs. 9 minutes for 200 epochs), and the counterfactual VAE training took less than 5 minutes.

\subsection{Further Results on Performance Evaluation}
We present the performance of \model{} and other baselines on downstream classification tasks by setting gender as the sensitive attribute in the main manuscript. We here present the extra results on race to support \model{}'s generalizability on different sensitive attributes. In the race attribute, UCI Adult, COMPAS, and LSAC have five, five, and six classes, respectively. UCI German Credit dataset does not include race information and hence is skipped. The evaluation results in Table~\ref{table:overallresults_race} demonstrate that \model{} learns data distribution of critical features while minimizing spurious information from the multi-class sensitive attribute.

\begin{figure*}[t!]
\centering
\begin{subfigure}[t]{0.23\textwidth}
\captionsetup{justification=centering}
       \centering\includegraphics[height=3.2cm]{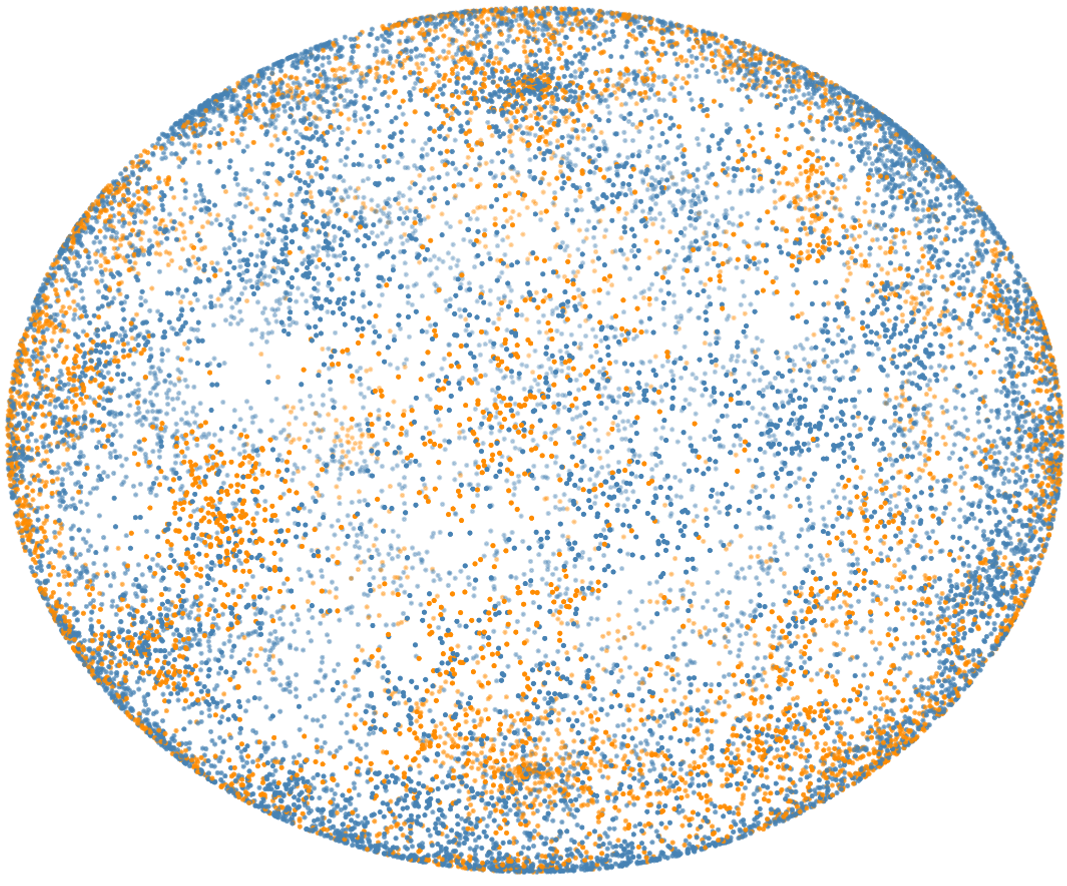}
      \caption{SCARF \\(AUC=0.92)}
      \label{fig:scarf}
\end{subfigure}
\hspace{1mm}
\begin{subfigure}[t]{0.23\textwidth}
\captionsetup{justification=centering}
       \centering\includegraphics[height=3.2cm]{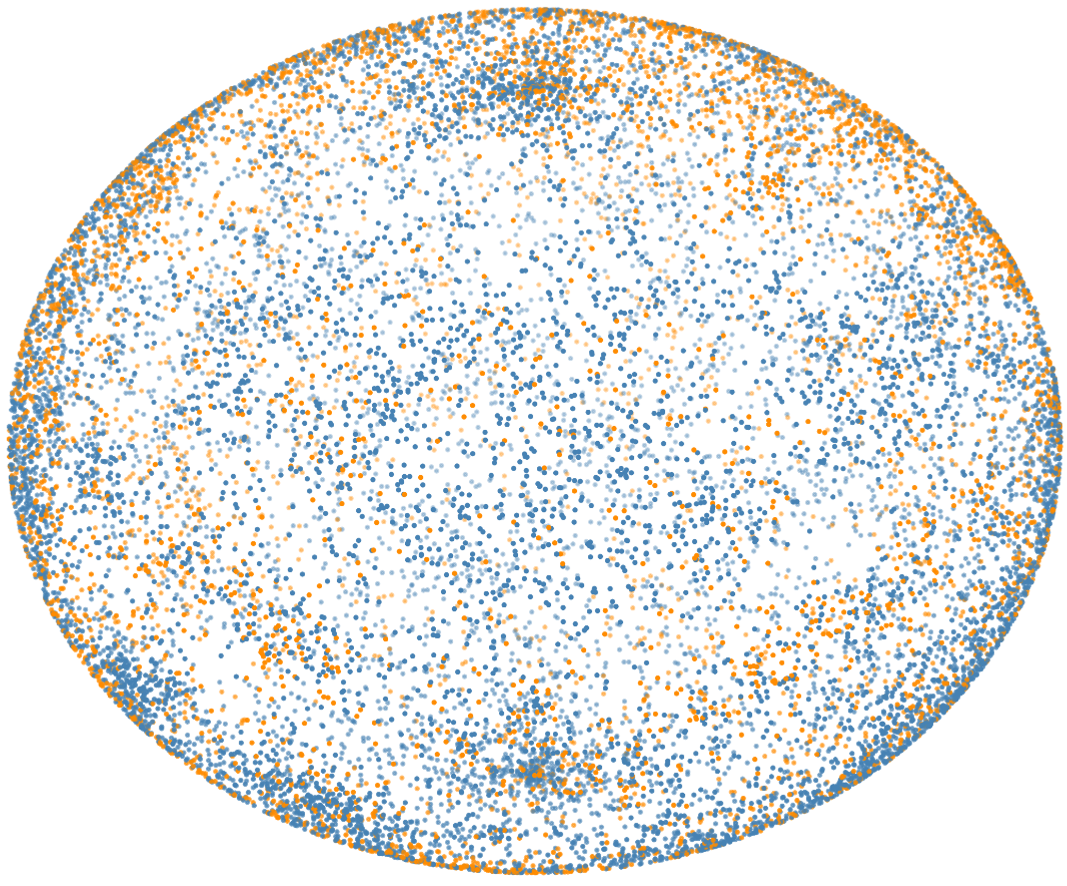}
      \caption{\model{} 1 epoch \\(AUC=0.75)}
      \label{fig:epoch1}
\end{subfigure}
\hspace{1mm}
\begin{subfigure}[t]{0.23\textwidth}
\captionsetup{justification=centering}
      \centering\includegraphics[height=3.2cm]{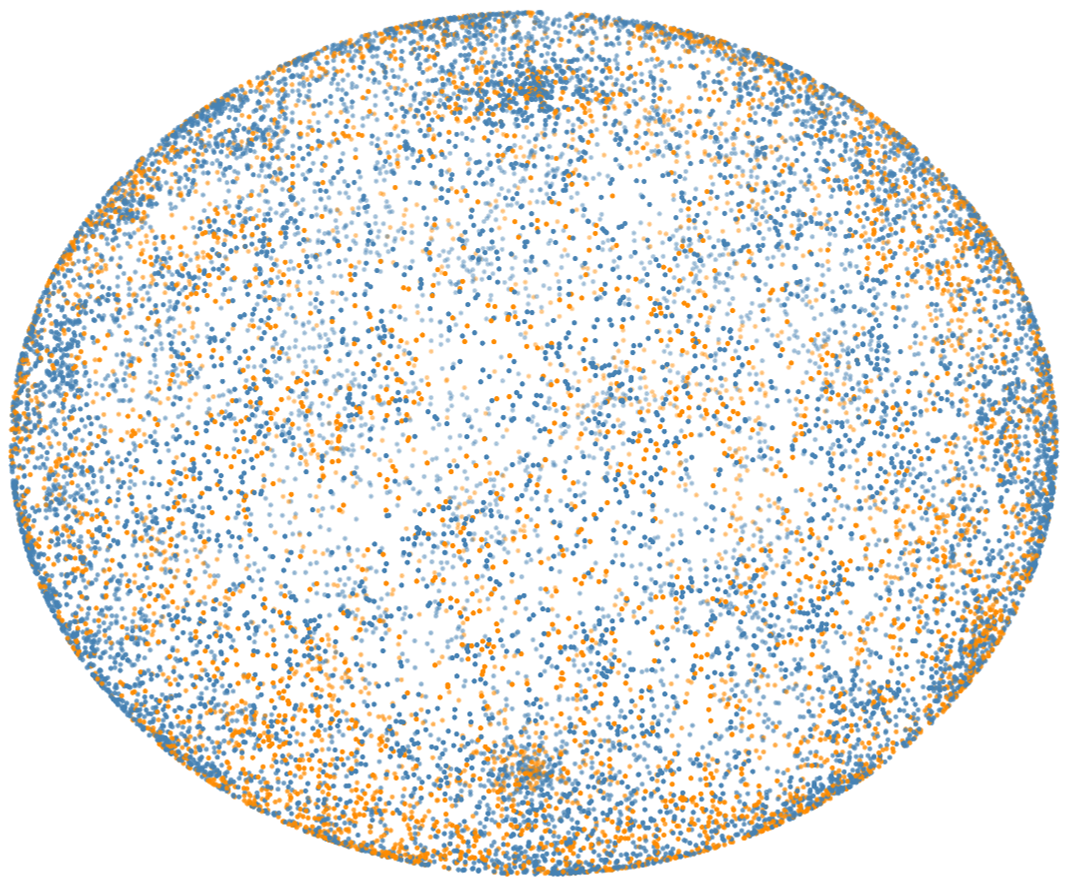}
      \caption{\model{} 10 epochs \\(AUC=0.72)}
      \label{fig:epoch10}
\end{subfigure}
\hspace{1mm}
\begin{subfigure}[t]{0.23\textwidth}
\captionsetup{justification=centering}
       \centering\includegraphics[height=3.2cm]{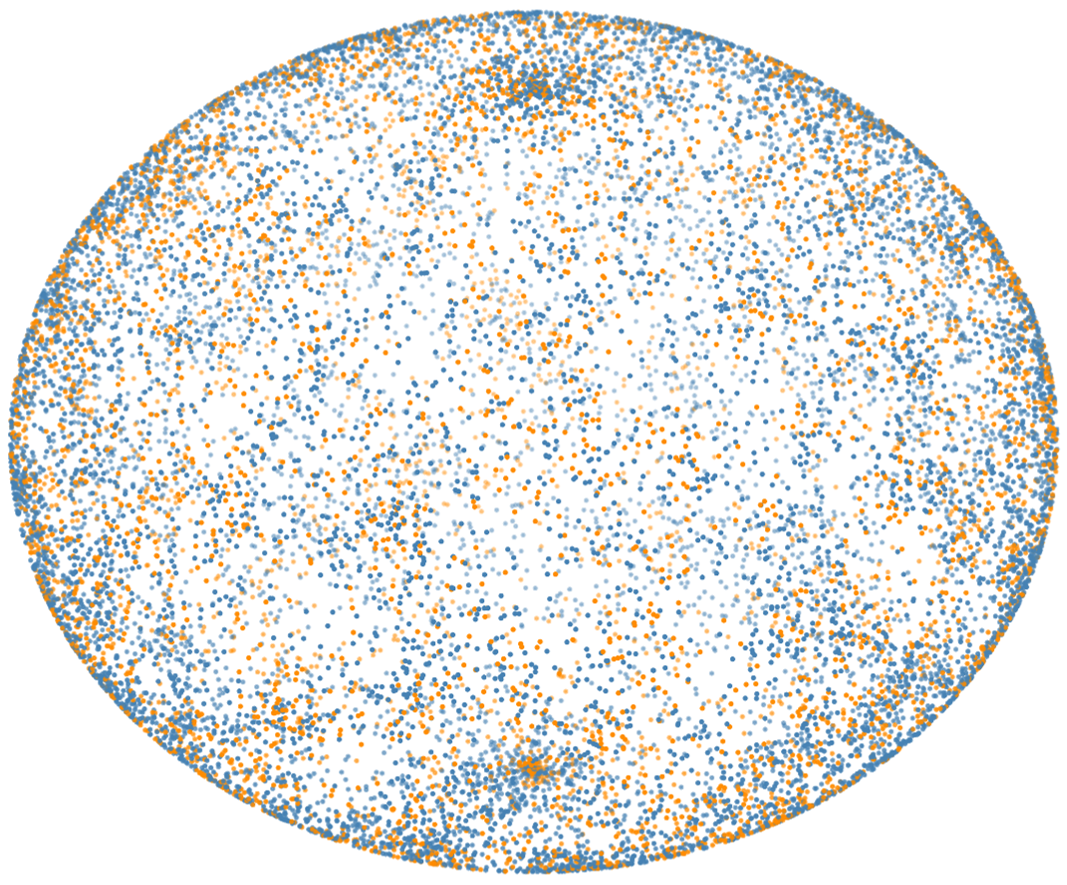}
      \caption{\model{} 200 epochs \\(AUC=0.61)}
      \label{fig:epoch200}
\end{subfigure}
\caption{UMAP visualization of trained embeddings over UCI Adult dataset from (a) SCARF and (b-d) \model{} with different epochs. Each dot represents the embedding of each individual, and color displays his/her gender. With these embeddings, we fit a classifier to predict the sensitive attribute (i.e., gender) of the test instances, anticipating that sensitive attribute cannot be determined from \model{} embeddings. As we expected, classifying the sensitive attribute using \model{} embeddings was difficult (i.e., AUC=0.61). We can infer that sensitive information was avoided in the process of creating the embedding.
}
\label{fig:qualitative}
\end{figure*}


\subsection{Further Results on Ablation study}
To confirm the effectiveness of the proposed counterfactual sample generator, we further assessed two ablations on loss objectives: (1) without cyclic consistency loss $L_{\text{cyc}}$ (Eq.~\ref{eq:cyc_loss}) and (2) without reconstruction loss in $L_{\text{vae}}$ (Eq.~\ref{eq:vae_loss}). Table~\ref{Tab:counterfactual-ablation} reports the test set performance of the logistic regression model, after fitting with synthetic counterfactual samples from each baseline. When we conduct experiments over four datasets, our model with full components outperforms baselines in terms of both the AUC and F1-score.

\begin{table}[!h]
\setlength{\tabcolsep}{2.5pt}
\centering
\caption{Ablation study of counterfactual sample generator. The proposed model with full components shows the best classification performance on the original data.}
\resizebox{\linewidth}{!}{
\begin{tabular}{lcccccccc}
\toprule
 & \multicolumn{2}{c}{Adult} & \multicolumn{2}{c}{Credit} & \multicolumn{2}{c}{Compas} & \multicolumn{2}{c}{LSAC} \\ \cmidrule{2-9}
Training set  & AUC & F1 & AUC & F1 & AUC & F1 & AUC & F1 \\ \midrule
Original          & 0.91 & 0.78 & 0.76 & 0.68 & 0.74 & 0.68 & 0.85 & 0.65 \\ 
Ours & 0.89 & 0.76 & 0.75 & 0.62 & 0.73 & 0.67 & 0.84 & 0.63 \\
w/o Cyclic loss ($L_{\text{cyc}}$) & 0.88 & 0.73 & 0.71 & 0.61 & 0.72 & 0.67 & 0.84 & 0.66 \\
w/o Reconstruction in ($L_{\text{vae}}$) & 0.35 & 0.45 & 0.35 & 0.45 & 0.48 & 0.42 & 0.36 & 0.47 \\
\bottomrule 
\end{tabular}}
\label{Tab:counterfactual-ablation}
\end{table}

\subsection{Further Results on Qualitative Analysis}
Figure~\ref{fig:qualitative} visualizes the embeddings over the UCI Adult dataset from four models: SCARF and \model{} at three different points of training (1, 10, and 200 epoch). SCARF does not consider any fairness requirements, and hence the learned representations of the same protected group (i.e., gender) are placed nearby, forming a local cluster in the embedding space (see Fig.~\ref{fig:scarf}). When we fit the logistic regression model to predict the sensitive attribute on top of the embeddings, the model reports very high AUC values. Meanwhile, the sensitive information is debiased across the epochs, and the sensitive attribute becomes indistinguishable in the embedding space (see Fig.~\ref{fig:epoch1}--\ref{fig:epoch200}). The AUC result in this new embedding gradually decreases with the training epoch.

\end{document}